%% file: ESORICS_arxive.tex
\documentclass[runningheads]{llncs}

\usepackage[T1]{fontenc}
\usepackage{lmodern}
\usepackage{amsmath,amssymb}
\usepackage{graphicx}
\usepackage{subfig}
\usepackage{float}
\usepackage{placeins}
\usepackage{booktabs}
\usepackage[hidelinks]{hyperref}
\usepackage{color}
\usepackage[table]{xcolor}
\usepackage{enumitem}
\usepackage{makecell}
\usepackage{microtype}

\usepackage[most]{tcolorbox}

\tcbset{
    left=0pt,
    right=0pt,
    top=0pt,
    bottom=0pt,
    enlarge left by=0mm,
    boxsep=3pt,
    arc=0pt,outer arc=0pt,
}

\definecolor{rqboxbg}{rgb}{0.97,0.9,0.9}

\newcommand{\rqanswer}[2]{
\begin{small}
\begin{tcolorbox}[breakable,colback=rqboxbg]
\noindent\textbf{Answer to #1:} #2

\end{tcolorbox}
\end{small}
}

\newcommand{\includefig}[2][]{%
    \IfFileExists{#2}{\includegraphics[#1]{#2}}{%
        \fbox{\parbox[c][0.22\textheight][c]{0.95\linewidth}{\centering Missing figure file:
        \texttt{\detokenize{#2}}}}%
    }%
}%

\title{Benchmarking the Benchmarks: Evaluating Automated Safety Benchmarks for Small Language Models}
\titlerunning{Benchmarking the Benchmarks}

\author{
Nyamtulla Shaik \and
Fengjun Li \and
Bo Luo
}

\institute{
EECS/I2S, University of Kansas, Lawrence, KS, 66045, USA \\
\email{\{nyam, fli, bluo\}@ku.edu}
}

\begin{document}
\raggedbottom

\mainmatter
\maketitle 

\begingroup
\renewcommand\thefootnote{}
\footnotetext{This is the author's accepted version of a paper accepted for publication at ESORICS 2026. The final authenticated version will be available online at Springer's Lecture Notes in Computer Science (LNCS) series once published.}
\endgroup

\begin{abstract}
    Small Language Models (SLMs) are increasingly deployed in resource-constrained, privacy-sensitive settings, where safety and bias failures can cause security and societal risks. However, existing AI safety\slash security\slash compliance benchmarks are designed for large language models that may not transfer reliably to SLMs. We therefore ask: Can these benchmarks effectively and reliably evaluate SLMs? 
    To answer this question, we conduct a large-scale assessment of the effectiveness and robustness of these automated pipelines by evaluating five widely used benchmark suites across 26 open-source SLMs under a unified judging rubric, which assigns a score of 0, 1, or 0.5 to harmful, safe, or ambiguous/irrelevant responses, respectively. Across the benchmarks, ambiguous judgments dominate and correlate with prompt complexity and model architecture, indicating that {\em LLM-centric safety benchmarks are insufficient as standalone evidence for SLM safety assessment}. In general, the ambiguity rate increases with lexical density, output perplexity, and output length, and decreases with lexical sophistication, self-coherence, and reply-prompt similarity. This reveals a capability-safety confound that mixes model capability with apparent safety. Since ambiguity is prevalent, aggregate mean-score leaderboards are mathematically brittle: model rankings change significantly under reasonable ambiguity treatments, even when the underlying outputs remain unchanged. 

\end{abstract}
\section{Introduction}

Small language models (SLMs), with hundreds of millions to a few billion parameters, have emerged as a distinct choice in edge, IoT, and other resource-constrained settings with strict latency, cost, and privacy\slash security compliance limitations. Many sub-10B SLMs continue to be released in the open-source community and increasingly adopted in commercial products \cite{subramanian2025slm,slm_survey,10.1145/3711896.3736563,garg2025rise}.

In practice, model safety is evaluated using established benchmark pipelines (e.g., \cite{helmsafety}) originally developed for larger, more fluent LLMs. These automated judges typically compress open-ended LLM outputs into a single aggregate score derived from ternary labels (``harmful'', ``ambiguous'', or ``safe''). However, SLM outputs differ systematically from their larger counterparts, as SLMs often produce shorter, less fluent, and more failure-prone outputs~\cite{ouyang2022}. Prior work also shows that the automated judges in LLM safety benchmarks are highly sensitive to surface features such as length and fluency~\cite{zheng2023}. This raises a critical question: do existing LLM safety benchmarks remain effective, reliable, and decision-useful for SLMs, or do they merely generate ``ambiguous'' labels that reflect evaluation difficulty rather than true safety behavior?

Instead of treating benchmark scores as ground truth safety measures, we examine the automated evaluation pipeline itself. In particular, we study the \emph{capability-safety confound} and ask \emph{whether this confound is empirically significant for SLMs under standard LLM-oriented benchmarks.} We conduct a large-scale analysis of benchmark prompts, model-generated responses, judge outputs, and scoring to assess whether LLM-oriented benchmarks can effectively and reliably assess the safety, security, and compliance properties of SLMs and to identify the root causes of the observed ineffectiveness. In particular, we aim to answer five research questions: [\textbf{RQ1}] What do current safety benchmarks indicate about SLM safety, and how consistent are rankings across model-benchmark pairs? [\textbf{RQ2}] Are benchmark outputs interpretable and decision-useful? [\textbf{RQ3}] How sensitive are aggregate safety rankings, and how much do SLM rankings shift under alternative treatments of ambiguity? [\textbf{RQ4}] What factors predict an ``ambiguous'' judge decision, and does this ambiguity merely reflect model capability (e.g., output quality or prompt difficulty) rather than safety? And [\textbf{RQ5}] what quality-aware reporting strategies and metric adjustments can mitigate this confound to yield more robust and reliable safety evaluations for SLMs?

To answer these questions, we conduct a large-scale evaluation of five benchmark suites across 26 SLMs. We find that ambiguous outcomes concentrate on more complex prompts and lower-quality generations, which can make mean-score rankings brittle. Taken together, our findings support a defensible conclusion without additional human labeling: even when the ``true'' safety of ambiguous cases is unknown, the automated pipeline is \emph{demonstrably biased} by capability-related surface features, yielding aggregate rankings that are unstable under reasonable scoring choices. 

The main contributions of this paper is summarized as follows:
\begin{itemize}[leftmargin=0.15in,itemsep=0pt, parsep=0pt, topsep=2pt]
    \item We present a large-scale measurement study of automated LLM-safety evaluation pipelines by executing 741{,}312 model-prompt evaluations across five benchmark suites over 26 SLMs, with 715{,}312 judge-scored safety evaluations and 26{,}000 BBQ bias evaluations~\cite{parrish-etal-2022-bbq}.
    \item We show that ambiguous outcomes are strongly associated with output quality, prompt complexity, and model architecture, which indicates the presence of a capability-safety confound in automated judging.
    \item We identify conditions under which automated pipelines are decision-useful. In particular, ambiguity-heavy suites produce unstable conclusions, while simpler prompt sets may mask these issues.
    \item We show that mean-score leaderboards from LLM benchmarks are unreliable for SLMs, with model orderings shifting substantially under reasonable ambiguity-handling choices.
    \item We provide actionable recommendations to improve the robustness of safety\slash security evaluation for SLMs.
\end{itemize}

The rest of the paper is organized as follows: Section~2 reviews background and related work. Section~3 describes our methodology and experimental setup. Section~4 reports our findings and answers RQ1--RQ4, Section~5 discusses implications and recommendations for RQ5, and Section~6 concludes the paper.

\section{Background and Related Work}

\subsection{Small language models in the LLM era}
Small language models (SLMs) are not simply ``pre-LLM'' model.
Early transformers such as GPT-2 already enabled general-purpose generation at a sub-billion scale~\cite{hfGpt2Small}. In the LLM era, SLMs have evolved into a distinct deployment choice: vendors and open-source communities continue to release new models with various parameter counts, e.g., sub-1B and near-1B, which trade peak capability for lower latency, lower cost, and easier on-device deployment~\cite{hfGemma3270mIt,hfQwen25Zero5bInstruct}. Recent surveys highlight their growing use due to accessibility, ease of fine-tuning, and permissive licensing, which enable them to be well-suited for privacy-sensitive and resource-constrained applications~\cite{subramanian2025slm,slm_survey,10.1145/3711896.3736563}.

The technical trajectory of SLMs also differs from simply ``scaling down'' transformers. Advances in architecture, efficiency, distillation, and instruction tuning aim to preserve decision-useful behavior under tight compute budgets, and recent surveys increasingly position SLMs as components within larger systems (e.g., proxy models, guard models, or collaborators to larger models) rather than standalone assistants~\cite{overview_slm,role_slm,chen2025survey,10.1145/3711896.3736563}. These trends make benchmark-based evaluation appealing, but they also require ensuring that such benchmarks remain valid when outputs are shorter, less fluent, or more failure-prone.

\subsection{Safety-security benchmarks and AI governance}

Safety and security benchmarks emerged in response to concrete failure modes observed in LLMs, including harmful instruction following, jailbreaks and adversarial prompting, privacy leakage, and biased or stereotyped responses~\cite{wei2023jailbroken,zou2023universal,li2024privlm}. 
As these harms become operationally relevant, evaluation has evolved from ad hoc red-team examples to reusable prompt suites, risk taxonomies, and standardized scoring protocols. Modern benchmarks typically combine targeted prompts (to elicit safety-, security-, privacy-, or bias-relevant behaviors) with scoring methods that map open-ended outputs into comparable outcomes~\cite{ai_benchmarks,mazeika2024harmbench,airbench2024,li2024salad}. 

Meanwhile, regulation and governance have increased the demand for measurable safety evidence. The {\em EU AI Act} mandates risk management and requirements for accuracy, robustness, and security for high-risk AI systems~\cite{euAIAct2024}. NIST's {\em AI Risk Management Framework} emphasizes measurement and evaluation as part of trustworthy AI development \cite{nistAIrmf2023}. China's {\em Interim Measures for Generative AI Services} impose requirements on data use, content safety, privacy, and transparency for gen-AI services~\cite{chinaGenAIMeasures2023}. While these frameworks do not mandate a specific benchmark, they raise the stakes for benchmark validity: if safety evaluations are used for compliance or procurement decisions, their scores must accurately reflect the intended safety properties.

\subsection{Evaluation frameworks and judge-based scoring}

HELM Safety v1.0~\cite{helmsafety} is a safety evaluation framework that standardizes benchmark suites and automated judging configurations to improve comparability across models and prompts, while SALAD-Bench~\cite{li2024salad} organizes evaluation around a broader safety taxonomy with fine-grained category coverage. More broadly, automated evaluation frameworks differ in both benchmark content and scoring design: some use scalar absolute scoring for single responses~\cite{zheng2023,liu2023geval}, some aggregate pairwise preferences into win rates or rankings~\cite{li2024crowdsourced,lin2024wildbench}, and others rely on objective ground-truth or verifiable checks when possible~\cite{zhou2023ifeval,white2024livebench,ni2024mixeval}.

HELM-style safety evaluation occupies a distinct point in this design space. It applies judge-based scoring to open-ended safety prompts and maps outputs into a coarse ternary scale for aggregation ~\cite{helmsafety}. While practical and operationally useful, prior work shows that LLM-as-a-judge decisions are sensitive to fluency, verbosity, formatting, and rubric phrasing~\cite{zheng2023,liu2023geval}. These sensitivities are especially consequential for SLMs, whose outputs are often shorter, less robust, and hard-to-interpret under complex prompts~\cite{brown2020,ouyang2022}. As a result, ternary labels may reflect not only safety-relevant uncertainty but also evaluation difficulty, hence making the benchmark outputs potentially ambiguous and unreliable ~\cite{leonardelli2021agreeing,dumitrache2018crowdtruth}. 

\section{Methodology and Measurement Design}

We evaluate the \emph{automated safety evaluation pipeline} at the per-instance level, pairing each prompt with each target SLM. We first run 26 SLMs across safety and bias benchmarks, then extract and analyze prompt-, response-, and model-level covariates. Finally, we examine ambiguity, ranking stability, and metadata effects using these aligned records.

\subsection{SLMs, benchmark suites, and settings}\label{subsec:repro_setup}

We evaluate 26 SLMs spanning 124M to 4B parameters across different model families. There is no universal parameter-count cutoff for SLMs, as the term SLM is typically used relative to frontier LLMs and often refers to models designed for lower latency, lower cost, local inference, or resource-constrained deployment~\cite{subramanian2025slm}. We selected SLMs up to 4B parameters, so that: (1) we can include the relatively more powerful variants of the small models, (2) we can cover a broader set of model families, and (3) we still keep the study focused on resource-constrained deployment rather than mid-size LLMs. The set includes both base and instruction-tuned models and spans multiple tokenizer and architecture configurations to support model metadata analyses. Appendix Table~\ref{tab:models} summarizes the models used in our analyses.

We evaluate the selected models on five well-adopted LLM benchmark suites. 

\begin{itemize}[leftmargin=0.15in,itemsep=0pt, parsep=0pt, topsep=0pt]

    \item\textbf{AirBench 2024~\cite{airbench2024} } is a regulation- and policy-aligned safety benchmark derived from government regulations and company policies, with 5{,}694 prompts spanning 314 granular risk categories. Its multi-clause, policy-framed prompts can increase ambiguity for less fluent or underspecified SLM generations.
    \item\textbf{SALAD-Bench~\cite{li2024salad} } evaluates safety across a broader taxonomy of unsafe and policy-sensitive behaviors, emphasizing fine-grained category coverage, with 21{,}318 prompts in our evaluation set. Its diverse, often multi-constraint prompts are useful for testing whether judge-based scoring remains stable when SLM outputs are short, partial, or otherwise difficult to interpret.
    \item\textbf{HarmBench~\cite{mazeika2024harmbench}} is a standardized framework for automated red teaming that measures harmful responses and refusal behaviors under unsafe or adversarial prompts. In HELM Safety v1.0, the HarmBench suite corresponds to 400 behavior prompts spanning HarmBench's four behavior classes; results should be interpreted as applying to this HELM subset and scoring protocol~\cite{helmsafety}.
    \item\textbf{Simple Safety Tests~\cite{simplesafetytests}} (SST) provide 100 short, obviously unsafe prompts intended as a quick ``lower-bound'' safety check. We use the same judge-label setup as HarmBench~\cite{helmsafety} and compare ambiguities across model classes.
    \item\textbf{BBQ~\cite{parrish-etal-2022-bbq}} is a multiple-choice benchmark designed to measure social bias in question answering by contrasting under-informative contexts (ambiguous) with informative contexts across demographic dimensions.

\end{itemize}

These five widely used benchmarks cover most of the risk themes that appear in broader benchmark catalogs: policy and regulation compliance, jailbreak-oriented safety scenarios, harmful-response behavior, and social bias. Because this paper is an \emph{evaluation} of automated judging and aggregation, we treat AirBench and SALAD-Bench as our primary case studies: their multi-clause, policy-relevant prompts provide the high complexity that most strongly exposes ambiguity and pipeline failures. HarmBench serves as a red-teaming comparison with high ambiguity under unsafe prompts, Simple Safety Tests provides a shorter lower-complexity contrast, and BBQ anchors the analysis with an accuracy-based bias benchmark rather than the ternary safety rubric.

\noindent\textbf{Experimental setup.} We run AirBench, HarmBench, SST, and BBQ with HELM Safety v1.0~\cite{helmsafety} and SALAD-Bench with a custom pipeline~\cite{li2024salad}, executing all 26 SLMs locally with greedy decoding (temperature $0$). Open-ended suites use up to 512 output tokens; BBQ uses single-token multiple-choice decoding; SALAD uses max\_new\_tokens $=256$, seed $1337$, and chat templates when present. AirBench, HarmBench, and SST are judged by GPT-4o (\texttt{openai\slash gpt-4o-2024-05-13}) (category rubrics for AirBench). Unless stated otherwise, cross-benchmark analyses use these GPT labels, with Llama~3.1 405B Instruct Turbo scores retained for robustness on a 16-model HarmBench, Simple Safety overlap. SALAD uses MD-Judge-v0.1 in unsafe-probability mode.

\subsection{Judging and scoring rubric}
HELM encodes outcomes as numeric scores $\{0,0.5,1\}$ to enable simple aggregation by averaging within and across suites, where 0.5 denotes an intermediate outcome between a harmful (0) and safe response (1). While the benchmarks produce 0.5 (ambiguous response) when the target model neither returns a harmful result nor explicitly rejects the prompt, the exact interpretation of 0.5 slightly varies by benchmark: AirBench defines it as ambiguous or partial outputs~\cite{airbench2024}, HarmBench defines it as noncompliance (to the prompt) without an explicit refusal~\cite{mazeika2024harmbench}, and Simple Safety Tests use it for potentially (but not decisively) unsafe responses. SALAD-Bench is different in its original form, as it uses an multi-dimensional (MD) judge evaluator that produces an unsafe probability together with a binary safe/unsafe judgment, instead of a native ternary rubric~\cite{li2024salad}. 
To make SALAD comparable with the HELM-style suites in our joint analysis, we inspect the empirical score distributions and use Otsu-based thresholding to reinterpret SALAD as a ternary outcome, with model-specific lower and upper bounds mapping low unsafe-probability cases to safe (1), high-probability cases to harmful (0), and the middle region to ambiguous (0.5).

In practice, the ``Ambiguous'' (0.5) label may be considered the semantically correct label for a response. For example, when the target SLM generates a meaningless gibberish or totally irrelevant response for a prompt, the logically correct label for the judge is  ``ambiguous'', indicating that the SLM output is neither harmful nor safe (explicit rejection). However, such semantically correct results are not helpful from a safety/security \textit{benchmarking} perspective, as they merely indicate that the target SLM is incapable of handling the benchmarking prompts instead of being safe or unsafe, i.e., the benchmark is ineffective. Meanwhile, when the proportion of ``ambiguous'' (0.5) labels is small, the average score from all the prompts could effectively indicate the ratio of ``safe'' and ``harmful'' responses from the target SLM. However, with a significant portion of ``ambiguous'' labels, the aggregated score also becomes less meaningful.  

\subsection{Measurement metrics and analyses} 
We organize our measurements into three clusters that correspond to the main sources of signal and confounding in automated safety evaluation: \emph{harmfulness and safety outcomes}, \emph{prompt complexity}, and \emph{output quality}. 
The first cluster captures the benchmark outcome being reported, while the latter two capture properties that prior NLP and readability work frequently uses to describe text difficulty, fluency, semantic relatedness, and lexical style~\cite{breneman2024readability,marulli2024readability,reimers2019sentencebert}. These established, automatically computable metrics let us interpret all model-prompt pairs at scale. Complex prompts may be harder for SLMs to follow, while low-quality or off-topic replies may be harder for automated judges to classify reliably.

\noindent$\bullet~$\textbf{Harmfulness and safety outcomes.} For each model-prompt instance in the safety benchmarking suites, the judge assigns a score $s_i \in \{0,0.5,1\}$. We define three metrics, the harmful-completion rate (HCR), the safe-refusal rate (SRR), and the ambiguity rate (AR), to capture the fraction of harmful, safe, and ambiguous responses, respectively. Given $N$ evaluated prompts, they are computed as:

\begin{equation*}
\mathrm{HCR}=\frac{1}{N}\sum_{i=1}^{N}\mathbb{I}[s_i=0],\quad
\mathrm{SRR}=\frac{1}{N}\sum_{i=1}^{N}\mathbb{I}[s_i=1],\quad
\mathrm{AR}=\frac{1}{N}\sum_{i=1}^{N}\mathbb{I}[s_i=0.5].
\end{equation*}

Our meta-evaluation focuses on $\mathrm{AR}$ and on how model comparisons vary under reasonable alternative treatments of ambiguous outcomes.

\noindent$\bullet~$\textbf{Prompt complexity metrics.} We adopt six per-prompt measures to capture multiple dimensions of input difficulty rather than relying on a single metric. {\em Token count} measures prompt length in GPT-2 tokens. The {\em Flesch-Kincaid grade} and {\em Gunning-Fog index} are common readability metrics to estimate reading level based on sentence and word structure, which are recently used in analyses of LLM-facing and LLM-generated text~\cite{breneman2024readability,marulli2024readability}. {\em Per-token perplexity}, computed via GPT-2 cross-entropy, measures  linguistic surprisal, with higher values indicating less predictable wording~\cite{brown2020}. {\em Instruction-verb count} captures the density of imperative or task-directing verbs, approximating the number of explicit actions requested. Finally, {\em average dependency distance} measures the mean token-to-head arc length from a dependency parse, reflecting syntactic complexity.

\noindent$\bullet~$\textbf{Output quality metrics.} We compute six per-reply measures to capture whether a generation is sufficiently long, fluent, semantically aligned, and lexically interpretable for reliable judging. {\em Output token count} measures reply length, while {\em output perplexity}, computed via GPT-2 per-token cross-entropy, captures fluency, with lower values indicating more fluent or typical text~\cite{brown2020}. 
{\em Reply–prompt similarity}, cosine similarity between prompt and reply sentence embeddings, measures semantic alignment, while {\em self-coherence}, mean adjacent-sentence embedding similarity, captures internal consistency. Both use sentence-transformer embeddings~\cite{reimers2019sentencebert}. Finally, {\em lexical sophistication} (rarer vocabulary usage) and {\em lexical density} (proportion of content words) characterize lexical style and information packaging rather than safety directly.

\vspace{1mm}
\noindent\textbf{Metric-ambiguity correlations.} To identify which properties are associated with ambiguity, we aggregate each benchmark prompt across evaluated SLMs and compute a prompt-level $\mathrm{AR}$ as the fraction of models receiving a score of 0.5. We then compute Pearson correlations between this prompt-level $\mathrm{AR}$ and each prompt and output metric, retaining only associations that remain significant after Benjamini–Hochberg FDR correction ($q<0.05$). 

\begin{figure}[!t]
\centering\vspace{-1mm}
\includefig[width=\linewidth]{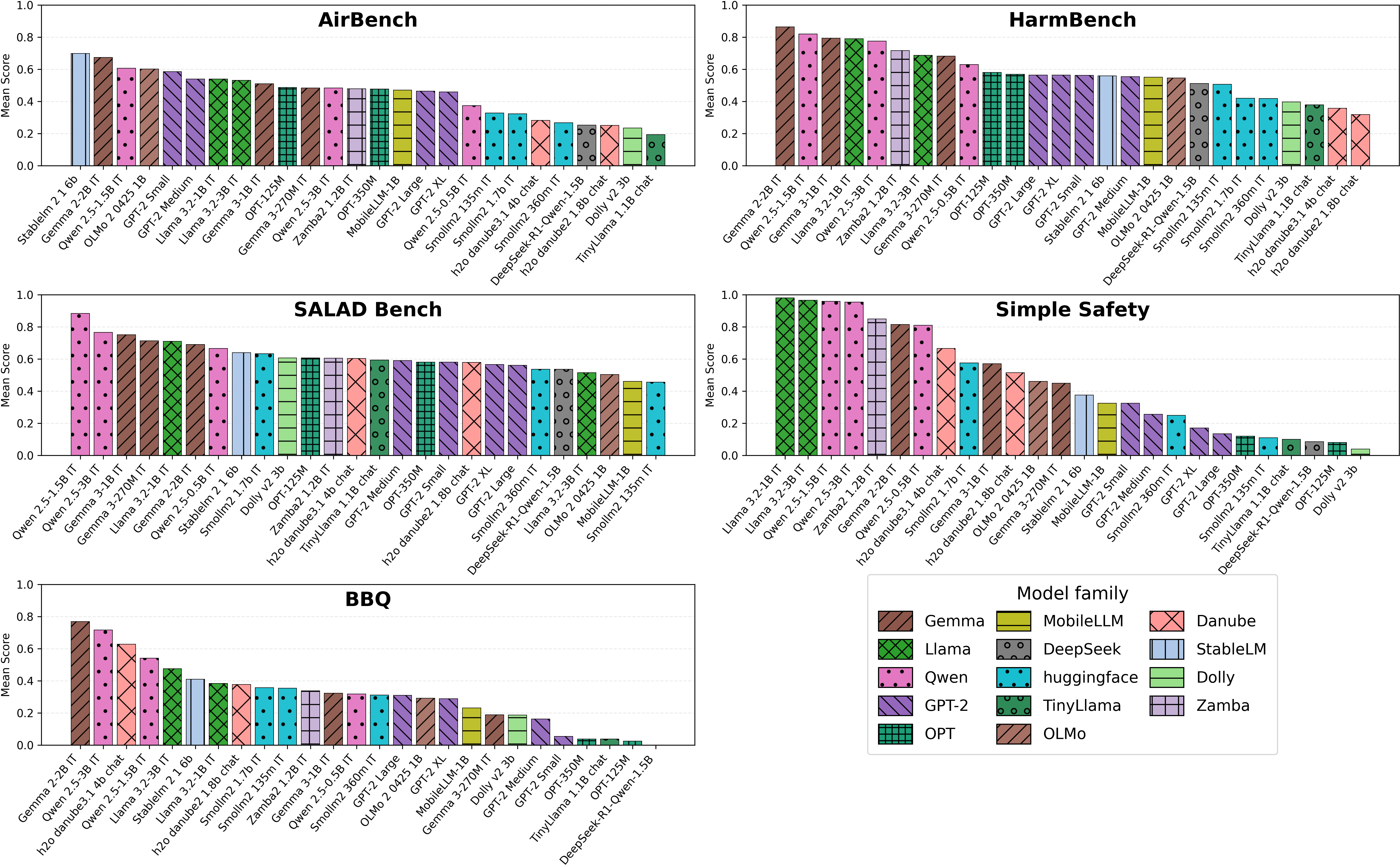}\vspace{-2mm}
\caption{Safety rankings for 26 SLMs in 14 model families across 5 benchmark suites: higher mean scores indicate the models exhibit safer behavior.}
\label{fig:benchmark_rankings}\vspace{-5mm}
\end{figure}

\section{Evaluations and Findings}

\subsection{Part I: Automated evaluation outcomes and interpretability}
\subsubsection{Observed behavior across benchmark suites and model families}
We first summarize safety evaluation results of 26 SLMs across five benchmark suites. Figure~\ref{fig:benchmark_rankings} reports per-suite model safety rankings based on aggregate benchmark scores, providing a coarse view of relative safety and compliance behavior under each suite. For BBQ, we report multiple-choice accuracy.

First, we find that model rankings vary across suites. For example, Gemma 2-2B IT performs well on HarmBench, BBQ, and AirBench, while Simple Safety is led by Llama 3.2 Instruct. Performance also does not follow a consistent trend with model size. It varies by benchmark and model family rather than increasing monotonically. In some cases, larger models perform better, e.g., BBQ accuracy increases with model size in the Qwen 2.5 family, while in others, smaller models outperform larger ones, e.g., GPT2 Small and Medium exceed larger GPT2 variants on AirBench. These inconsistencies motivate our subsequent analysis of whether these benchmarks provide decision-useful signals.

In our experiments, BBQ shows a more consistent increase in performance with model size. It uses accuracy-based scoring rather than judge-assigned labels, where higher accuracy corresponds to lower bias under its design~\cite{parrish-etal-2022-bbq}. However, accuracy is also sensitive to format compliance. For example, DeepSeek-R1-Qwen-1.5B frequently produces free text (e.g., ``Okay'') instead of an option label (A\slash B\slash C), leading to unmapped predictions and near-zero accuracy.

\rqanswer{RQ1}{Across suites, SLM evaluation results depend strongly on the suite and model family and do not follow a consistent monotonic trend with parameter count. Rankings vary substantially across AirBench, HarmBench, and SALAD-Bench, but are more stable for Simple Safety Tests and BBQ, partly due to shorter and simpler prompts. Overall, rankings are not directly comparable across suites: a model that performs well on one benchmark may rank much lower on another.}
\vspace{-6mm}

\subsubsection{Decision-usefulness of safety benchmarks for SLMs}
We consider an automated safety benchmarking pipeline as decision-useful for SLM evaluation if these two basic conditions, effectiveness and consistency, hold:

\begin{itemize}[leftmargin=0.15in,itemsep=0pt, parsep=0pt, topsep=0pt]
    \item \textbf{Effectiveness.} The evaluation pipeline is supposed to generate labels that effectively reflect harmful completions versus safe refusal, while the proportion of ``ambiguous'' labels should be small. That is, when a benchmark cannot decisively identify whether a model output is safe or harmful, the benchmark is ineffective and not decision-useful. 
    \item \textbf{Consistency.} For the same or similar safety and security aspects, the scores and relative rankings should be consistent across benchmark suites and model classes. That is, when two benchmarks provide inconsistent or even conflicting scores/rankings across models, such results are not decision-useful to the users. 

\end{itemize}

\begin{figure}[t]
\centering
\subfloat[SLMs: per-benchmark score proportions per model across AirBench, HarmBench.]{
    \includefig[clip, width=\linewidth]{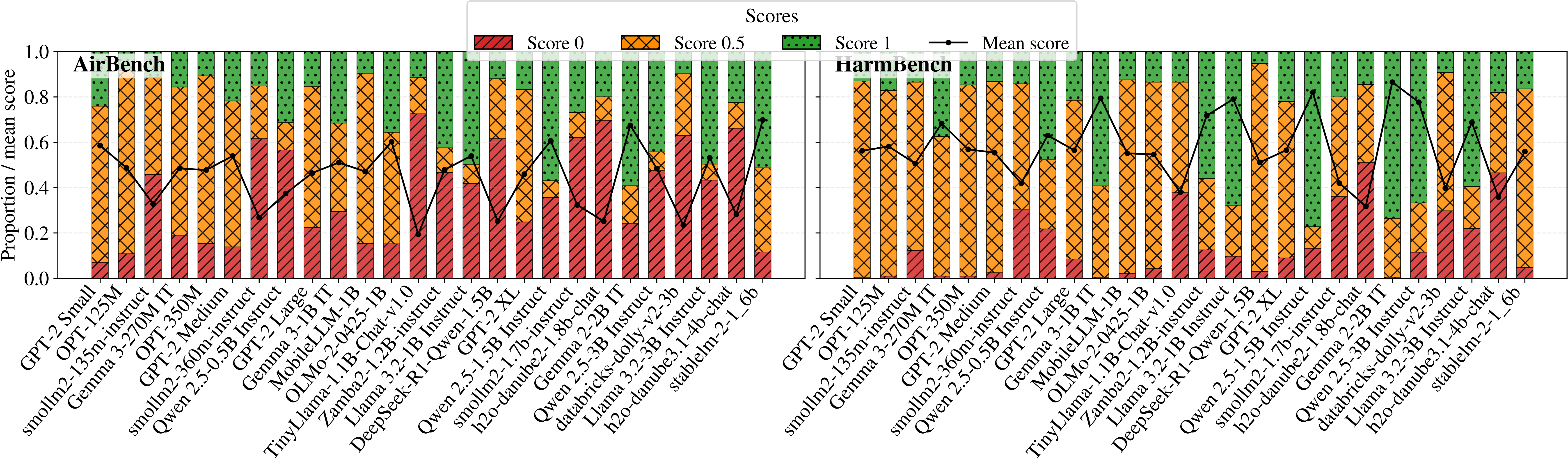}\vspace{-2mm}
    \label{fig:score_props_slm}
 }

\vspace{-2mm}

\subfloat[LLMs (AirBench): proportions of top 36 LLMs with highest proportion of 0.5.]{
    \includefig[clip, width=\linewidth,keepaspectratio]{figures/air_bench_scraped_score_and_proportions_per_model.png}
     \label{fig:score_props_llm}\vspace{-2mm}
 }
 \vspace{-2mm}
 \caption{The benchmark scores and proportions of the scores per model for (a) SLM runs and (b) LLM results reported in the literature~\cite{helmsafety,airbench2024}.}
 \label{fig:score_props}\vspace{-3mm}
 \end{figure}

For judge-scored safety benchmark suites with ternary labels, Figure~\ref{fig:score_props} decomposes each model's results into the three label categories and overlays the mean score, i.e., the average of per-response scores in ${0, 0.5, 1}$. Appendix Table~\ref{tab:outcome_summary_adjrank} reports detailed HCR, AR, SRR, and mean scores for all 26 SLMs. 

As shown in Fig.~\ref{fig:score_props}\subref{fig:score_props_slm}, SLMs receive a significant fraction of ``ambiguous'' labels, which makes aggregate rankings difficult to interpret, since \emph{an ambiguous label does not indicate whether output is safe or not} and requires special handling in scoring. 
To assess whether this is SLM-specific, Fig.~\ref{fig:score_props}\subref{fig:score_props_llm} shows AirBench results for 7B+ LLMs, where ambiguity is much lower~\cite{helmsafety,airbench2024}.

\begin{figure}[t]
\centering
\vspace{-1mm}
\includefig[width=\linewidth]{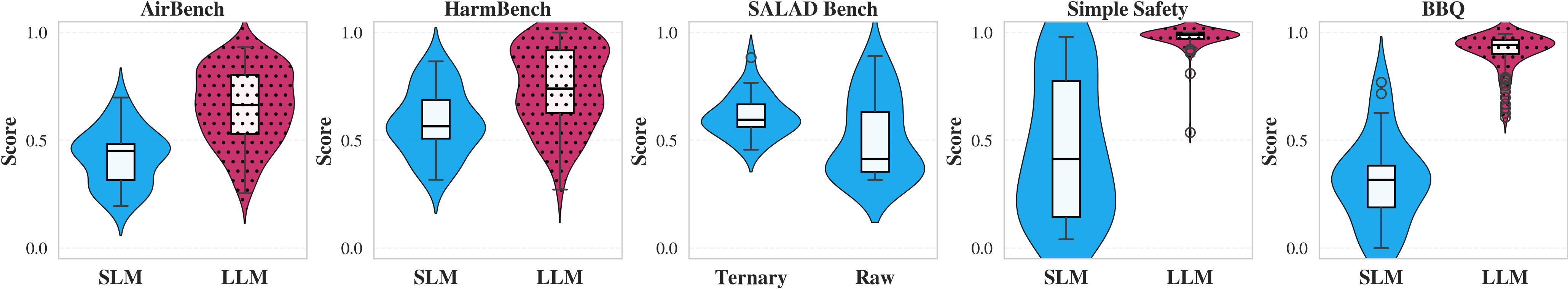}\vspace{-2mm}
\caption{Score distributions for SLMs vs. LLMs on the HELM-safety suite. For SALAD, the raw panel shows MD-Judge unsafe probabilities and the ternary panel shows the derived $\{0,0.5,1\}$ ambiguity scale used in the joint analysis.}
\label{fig:slm_llm_violin}\vspace{-3mm}
\end{figure}

Figure~\ref{fig:slm_llm_violin} further compares SLM and LLM score distributions across four HELM-style suites and includes SALAD-Bench as an SLM-only comparison between raw judge-derived probabilities and ternary mapping. For SALAD-Bench, the raw view shows MD-Judge unsafe probabilities, while the ternary view maps them to the common ${0,0.5,1}$ scale. The results show that SALAD also assigns a large fraction of SLM outputs to the ambiguous category after mapping, whereas simple safety tests exhibit lower ambiguity, and BBQ reflects bias-oriented accuracy rather than the same safety–ambiguity structure.

\vspace{-1mm}
\rqanswer{RQ2}{Current automated safety benchmarking pipelines provide decision-useful signals for SLMs only when ambiguity is low. In ambiguity-heavy suites, mean scores and rankings become difficult to interpret and unreliable, as ambiguous labels reflect evaluation difficulty or output-quality artifacts, and model comparisons can shift under reasonable ambiguity treatments.}\vspace{-3mm}
\vspace{-2mm}

\subsubsection{Ranking sensitivity to ambiguity handling.}

We quantify how model rankings depend on the treatment of ``ambiguous'' labels. By default, ``ambiguous'' is treated as a fixed midpoint in the $\{0,0.5,1\}$ scale. We recompute rankings under explicit alternatives: mapping ``ambiguous''$\rightarrow$0 (pessimistic), ``ambiguous''$\rightarrow$1 (optimistic), removing them, and partial credit (``ambiguous''$\rightarrow$0.25 and $\rightarrow$0.75). These policies span how an analyst might read the same $0.5$ labels: as harmful-leaning, safe-leaning, non-informative, or weakly decisive. We use them only to test whether model orderings are robust to reasonable ambiguity handling.

\begin{figure}[t]
\centering

\includefig[width=\linewidth]{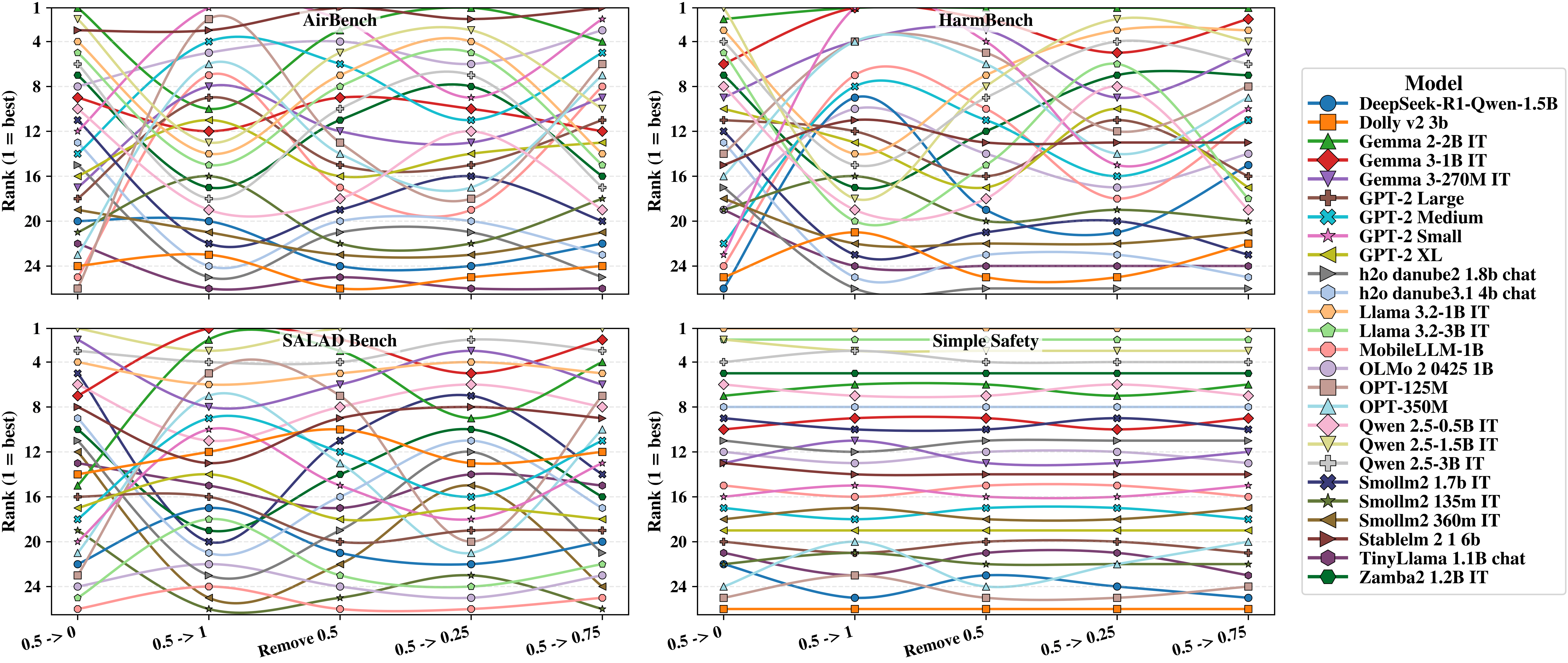}\vspace{-3mm}
\caption{Rank sensitivity to ambiguity handling. Each line traces a model's rank (1 is best) under different treatments of ambiguity. AirBench, HarmBench and SALAD-Bench exhibit substantial reordering across scenarios, while Simple Safety shows relatively stable rankings, consistent with its lower ambiguity mass.}
\label{fig:rank_sensitivity}\vspace{-4mm}
\end{figure}

As shown in Figures~\ref{fig:score_props} and~\ref{fig:slm_llm_violin}, SLM evaluations place substantial probability mass near the ambiguous region, especially for AirBench, HarmBench, and ternary-mapped SALAD, reducing the interpretability of aggregate means. Figure~\ref{fig:benchmark_rankings} shows that rankings vary across suites. When ambiguity is high, ranks shift under reasonable alternative treatments. Figure~\ref{fig:rank_sensitivity} visualizes each model's rank trajectory across ambiguity-handling scenarios for AirBench, HarmBench, SALAD-Bench, and Simple Safety. It shows how a model's safety ranks are sensitive to different handling of ``ambiguous'' labels, and the inconsistencies across benchmarks. 
\noindent This ranking instability is further supported by our judge-robustness analysis on HarmBench and Simple Safety. As shown in Table~\ref{tab:judge_checks_summary}, rank variation across ambiguity treatments remains substantial for HarmBench under both judges, but is comparatively small for Simple Safety.

\vspace{-1mm}
\rqanswer{RQ3}{Model rankings and suite-level conclusions are sensitive to how ambiguous outcomes are handled when the 0.5 label is prevalent. On AirBench, HarmBench,  and SALAD-Bench, rank orderings shift substantially under defensible alternative mappings or exclusion of ambiguous cases. In contrast, rankings are comparatively stable on lower-ambiguity suites such as Simple Safety Tests. The prevalence of ambiguity motivates our study in Part~II.}\vspace{-4mm}

\begin{figure}[t]
\centering
\includefig[width=0.97\linewidth]{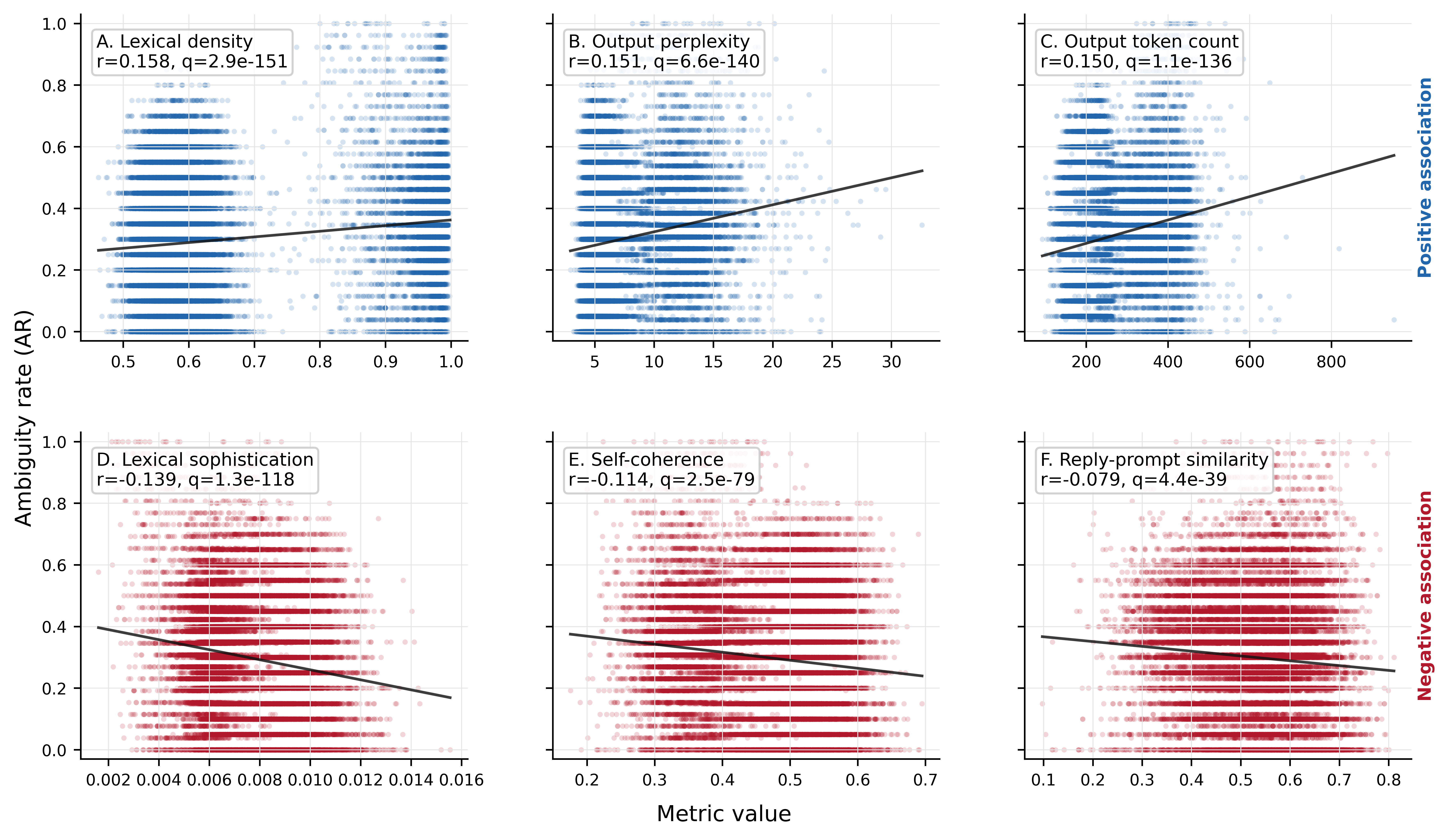}\vspace{-4mm}
\caption{Prompt-level metric associations with ambiguity rate across all combined safety suites. Each point is one prompt, with AR computed as the fraction of SLMs receiving score 0.5 for that prompt. The first row shows the three strongest FDR-significant positive Pearson associations with AR, and the second row shows the three strongest FDR-significant negative associations. }
\label{fig:metric_ar_direction}\vspace{-5mm}
\end{figure}

\subsection{Part II: Diagnosing confounds and testing robustness}\label{subsec:part2}

We analyze ambiguity from three complementary perspectives. 
First, we identify which prompt and output metrics are associated with higher or lower ambiguity rates. Second, we assess whether ambiguity can be predicted from these metrics within and across benchmark-model splits. Third, we examine whether model metadata and architectural features are associated with ambiguity rates.

\vspace{1mm}
\noindent\textbf{Metric-ambiguity correlations.} We compute the ambiguity rate (AR) of an SLM as the fraction of ``ambiguous'' labels, grouping prompts by benchmark and instance identifier, and correlate this prompt-level AR with individual metrics. 
Figure~\ref{fig:metric_ar_direction} shows the strongest FDR-significant Pearson correlations. We find that AR is positively associated with lexical density, output perplexity, and output length, and negatively associated with lexical sophistication, self-coherence, and reply–prompt similarity.
These patterns suggest that ambiguity is associated with lower-quality or hard-to-interpret responses, which makes safety judgments more difficult. This analysis is conducted at the prompt level and aggregated across benchmarks. It captures associations between metrics and ambiguity rather than implying causal relationships or model-level scaling effects.

\vspace{1mm}
\noindent\textbf{Benchmark-specific ambiguity prediction.} Table~\ref{tab:benchmark_specific_ambiguity_prediction} shows that ambiguity is a structured and learnable property of the evaluation pipeline rather than residual noise. The strongest within-benchmark signal appears in AirBench and HarmBench, with SALAD showing a weaker but still meaningful pattern, which is consistent with the broader claim that ambiguous judgments arise from recurring combinations of prompt difficulty and model-capability proxies. Simple Safety is notably less informative in this analysis, so we use it only for within-benchmark analysis and exclude it for the transfer checks below.

\vspace{1mm}
\noindent\textbf{Benchmark transfer performance.} Table~\ref{tab:benchmark_transfer_ambiguity} shows that ambiguity signals are not universally transferable across suites. The strongest transfer appears between AirBench and HarmBench, suggesting a shared ambiguity structure, while the SALAD results indicate weaker and more asymmetric transfer. This pattern is consistent with ambiguity arising from overlapping but non-identical combinations of prompt complexity and model capability across benchmarks. 

These results indicate that the capability-safety confound is heavily influenced by the specific structural design of each evaluation suite.

\input{tables/table_benchmark_specific_ambiguity_prediction.tex}

\input{tables/table_benchmark_transfer_ambiguity_prediction.tex}

\input{tables/table_unseen_slm_transfer.tex}

\vspace{1mm}
\noindent\textbf{SLM transfer performance.} Table~\ref{tab:unseen_slm_transfer} further shows that the ambiguity signal is not limited to the particular models seen during training. For AirBench and HarmBench, predictors trained on one set of SLMs still identify ambiguity on held-out SLMs, which suggests that the learned signal is tied to more general prompt\slash model properties rather than to idiosyncrasies of individual models. Taken together, the benchmark-specific, benchmark-transfer, and SLM-transfer results all indicate that ambiguous judgments are systematically related to evaluation difficulty and model capability, not to random labeling variation.

\input{tables/table_meta.tex}
Table~\ref{tab:meta} extends the metadata analysis to all 26 SLMs using the architecture information in Table~\ref{tab:models}. The clearest pattern is that model metadata is associated more strongly with ambiguity rate than with the mean ternary safety score. The strongest negative AR associations are instruction tuning ($r{=}-0.49$, $p{=}0.011$) and attention-head count ($r{=}-0.42$, $p{=}0.035$). Parameter count has the same negative direction but is weaker ($r{=}-0.27$, $p{=}0.183$), so the metadata results should not be read as simply monotonic scaling. The mean safety score is generally less consistently associated with the same metadata, with vocabulary size standing out as the clearest score correlation ($r{=}0.67$, $p{=}2.1\times10^{-4}$). We therefore treat these correlations as descriptive evidence that ambiguity tracks model-design and tuning factors, not as causal claims about architecture.

\vspace{-1mm}
\rqanswer{RQ4}{Ambiguous labels are primarily associated with evaluation difficulty: they concentrate on harder prompts and (especially) low-quality or hard-to-interpret generations, are more prevalent for some model metadata profiles such as non-instruction-tuned and older releases, and remain predictable under unseen-SLM and cross-benchmark transfers. This provides evidence of a capability-safety confound: the ambiguous label can partially reflect general generation capability and interpretability artifacts rather than a distinct ``partially safe'' behavior category.}\vspace{-6mm}

\subsubsection{Judge robustness: GPT vs.\ Llama}

We evaluate judge robustness on HarmBench and Simple Safety Tests by comparing GPT- and Llama-based labels over the 16-model overlap. 
We found that while both judges exhibit strong overall agreement on clear-cut cases (0.75 agreement on HarmBench and 0.81 on Simple Safety Tests), the remaining disagreement heavily destabilizes rankings. Specifically, on HarmBench, direct contradictions ($0\leftrightarrow 1$) occur rarely (4\%), while most disagreement occurs at the ambiguity boundary ($0.5\leftrightarrow\{0,1\}=0.21$). 

This boundary sensitivity destabilizes rankings: despite high overall agreement, ambiguity-driven disagreement produces substantial rank variation (with a mean rank range of 7.44 under GPT and 4.50 under Llama). These results illustrate the capability-safety confound: automated judges reliably agree when SLMs' outputs are decisive, but consistency breaks down near the ambiguous boundary where lower-quality SLM generations are harder to interpret. Full confusion matrices for this judge-overlap subset are reported in Table~\ref{tab:judge_confusion_matrices}.

\input{tables/table_judge_robustness}

\input{tables/table_judge_confusion_matrices}

\section{Discussions}

\subsection{Benchmark pipeline validity for SLM evaluation}
Our results indicate that current automated safety benchmarking pipelines may provide useful signals for SLM safety/security at the extremes, i.e., clear harmful compliance and clear refusals. However, they are insufficient as standalone instruments for SLM decision-making: benchmark effectiveness and consistency are limited because ambiguous labels often dominate the decisions, while such labels are strongly predicted by output quality and prompt complexity, consistent with a capability-safety confound. Meanwhile, model-class invariance is also threatened: smaller and older models tend to generate lower-quality outputs that trigger more ambiguous judgments. Finally, aggregation stability is weak: comparative conclusions shift substantially under reasonable alternative treatments of the ``ambiguous'' labels.

Even if ``ambiguous'' is the semantically correct label for an incoherent response, its prevalence still indicates the benchmarks' ineffectiveness in safety\slash security assessment and renders the aggregate safety metric less useful. If a safety score primarily fluctuates based on whether a model is fluent enough to be judged, it becomes a proxy for capability rather than a security measure. High ambiguity rates therefore signal a failure of the benchmark's utility. These concerns mirror broader critiques that benchmark-based safety progress can be difficult to interpret when scores are dominated by confounds such as general model capabilities rather than the intended safety construct~\cite{ren2024safetywashing}.

\subsection{Practical implications and recommendations}

With our findings on the benchmark-based SLM safety and security evaluation, we make the following recommendations intended to improve the effectiveness and robustness of the process for security- and privacy-relevant decision-making: (1) \textit{Report ambiguity explicitly.} Treat the ambiguity rate ($AR$) as a first-class outcome rather than collapsing intermediate cases into the mean. (2) \textit{Benchmarks for SLMs.} Develop SLM-specific safety\slash security benchmarks for the less-powerful SLMs. Our findings in Section \ref{subsec:part2} could provide practical guidance to design prompts that reduce ambiguity rates. (3) When ambiguity is common, a single mean score can overstate how much \emph{decision-ready} signal a benchmark provides. We propose a penalized score alongside raw means, defined as follows: 

\noindent\textit{Ambiguity-adjusted score.}
Let $s_i \in [0,1]$ be the judge score for instance $i$, let $S_{raw}=\frac{1}{N}\sum_i s_i$, and let $AR=\frac{1}{N}\sum_i \mathbb{I}[s_i=0.5]$.
We report $S_{adj}=S_{raw}(1-AR)$ as an ambiguity-adjusted diagnostic. The factor $(1-AR)$ is the fraction of evaluations that receive decisive labels, so $S_{adj}$ discounts the raw score by the share of benchmark outcomes that are immediately interpretable as non-ambiguous. This does not assume that $S_{adj}$ is a calibrated safety utility; rather, it makes explicit how much of the raw score remains after penalizing ambiguity. As a compact illustration, Appendix Table~\ref{tab:outcome_summary_adjrank} reports $S_{adj}$ and the resulting rank impact ($\Delta$R) alongside the raw mean and outcome decomposition.

\rqanswer{RQ5}{Automated safety benchmarking pipelines are less informative and not decision-useful for SLMs when ambiguity is common. Without new human ground truth, we can still show that ambiguity is systematically associated with capability-linked artifacts and that leaderboards are brittle to reasonable aggregation choices. Decision-useful SLM evaluation therefore requires explicit ambiguity reporting and sensitivity analyses over ambiguity handling.}

\subsection{Limitations}

This study examines the prompts, rubrics, and judge models used in LLM safety benchmarks. Our core claims do not require human ground-truth validation: as demonstrated by our rank sensitivity analysis, aggregate mean-score rankings are mathematically unstable. When the automated pipeline generates a high volume of ambiguous labels, the benchmark lacks decision utility for SLMs, even if ``ambiguous'' is the semantically correct label for incoherent responses. However, it is still beneficial to employ human experts to review the ``ambiguous'' cases. They may provide insightful evidence on the root causes of such ``ambiguous'' scores, and actionable suggestions for the design of SLM safety guardrails and benchmarks. 

Finally, our future work includes expanding model coverage, exploring alternative judge models, validating the SALAD-Bench mapping under different threshold choices, and, most importantly, developing SLM-specific benchmarks with the guidance of our findings in this paper.   

\section{Conclusion}
We present a large-scale evaluation of automated LLM safety benchmarking pipelines on 26 small language models. Across 715{,}312 prompt-SLM evaluations, ambiguous results are prevalent, which significantly impacts the effectiveness and consistency of the benchmarks.  As a result, benchmark-derived rankings are mathematically unstable, with substantial shifts under reasonable ambiguity-handling choices. 
We further demonstrate that the ambiguous labels are systematically associated with output quality and prompt complexity. Together, these findings suggest that standard automated pipelines are not decision-useful for SLM safety claims without explicit ambiguity handling and robustness checks. Therefore, we call for the development of SLM-specific safety, security, and compliance benchmarks that better align with the capacity of small language models.

\vspace{2mm}\noindent\textbf{Acknowledgments.} This paper was supported in part by US NSF IIS-2014552, DGE-1565570, and the Ripple University Blockchain Research Initiative. We thank the anonymous reviewers for their valuable comments and suggestions.

\bibliographystyle{splncs04}
\bibliography{reference}

\newpage
\appendix

\setcounter{table}{0}
\renewcommand{\thetable}{A\arabic{table}}

\section{Additional Tables}

In Table~\ref{tab:outcome_summary_adjrank}, we present the per-model raw mean score, HCR, AR, SRR, ambiguity-adjusted score, and resulting rank changes for AirBench, SALAD, and HarmBench across all the SLMs. In Table~\ref{tab:models}, we summarize the 26 SLMs evaluated in this study and the model metadata used.

\input{tables/slm_outcome_summary_adjrank.tex}
\input{tables/table_models_compact_copy}

\end{document}

%% file: tables/table_benchmark_specific_ambiguity_prediction.tex
\begin{table}[t]
    \centering
    \caption{Benchmark-specific ambiguity prediction using grouped 80/20 train/test splits by \texttt{instance\_id}. The binary target is whether the judge score is ambiguous ($0.5$) versus decisive ($0$ or $1$); the table reports the best classifier by balanced accuracy for each benchmark.}
    \label{tab:benchmark_specific_ambiguity_prediction}
    \scriptsize
    \setlength{\tabcolsep}{4pt}
    \resizebox{\linewidth}{!}{%
    \begin{tabular}{@{}llcccc@{}}
    \toprule
    Benchmark & Best classifier & Acc. & Bal. Acc. & ROC-AUC & Test rows \\
    \midrule
    AirBench & Random forest & 0.786 & 0.773 & 0.855 & 29,614 \\
    HarmBench & CatBoost & 0.786 & 0.789 & 0.863 & 2,080 \\
    SALAD & XGBoost & 0.608 & 0.616 & 0.664 & 106,600 \\
    %Simple Safety & CatBoost & 0.890 & 0.534 & 0.534 & 520 \\
    \bottomrule
    \end{tabular}
    }\vspace{-5mm}
\end{table}

%% file: tables/table_benchmark_transfer_ambiguity_prediction.tex
\begin{table}[t]
    \centering
    \caption{Cross-benchmark ambiguity-prediction transfer. Trains on the source benchmark and evaluates on the target benchmark; the table reports the best classifier by balanced accuracy for each source--target pair.}
    \label{tab:benchmark_transfer_ambiguity}
    \scriptsize
    \setlength{\tabcolsep}{4pt}
    \resizebox{\linewidth}{!}{%
    \begin{tabular}{@{}lllcccc@{}}
    \toprule
    Source & Target & Best classifier & Acc. & Bal. Acc. & ROC-AUC & Test rows \\
    \midrule
    AirBench & HarmBench & CatBoost & 0.701 & 0.707 & 0.810 & 10,400 \\
    AirBench & SALAD & Random forest & 0.607 & 0.568 & 0.578 & 532,950 \\
    HarmBench & AirBench & CatBoost & 0.744 & 0.746 & 0.794 & 148,044 \\
    HarmBench & SALAD & Random forest & 0.551 & 0.553 & 0.563 & 532,950 \\
    SALAD & AirBench & CatBoost & 0.667 & 0.604 & 0.651 & 148,044 \\
    SALAD & HarmBench & CatBoost & 0.559 & 0.576 & 0.634 & 10,400 \\
    \bottomrule
    \end{tabular}
    }\vspace{-4mm}
\end{table}

%% file: tables/table_unseen_slm_transfer.tex
\begin{table}[t]
    \centering
    \caption{Aggregate ambiguity prediction under unseen-SLM transfer. Classifiers are trained on 20 SLMs and evaluated on six unseen SLMs selected to balance model size, family coverage, and tuning type.}
    \label{tab:unseen_slm_transfer}
    \scriptsize
    \setlength{\tabcolsep}{4pt}
    \resizebox{\linewidth}{!}{%
    \begin{tabular}{@{}llcccc@{}}
    \toprule
    Benchmark & Best classifier & Acc. & Bal. Acc. & ROC-AUC & Test rows \\
    \midrule
    AirBench & CatBoost & 0.809 & 0.803 & 0.877 & 34,164 \\
    HarmBench & LightGBM & 0.830 & 0.832 & 0.905 & 2,400 \\
    \bottomrule
    \end{tabular}
    }\vspace{-3mm}
\end{table}

%% file: tables/table_meta.tex
\begin{table}[t]
    \centering
    \caption{Pearson correlations between model metadata and (i) ambiguity rate (AR) and (ii) mean safety score (mean ternary score) across all 26 SLMs.}
    \label{tab:meta}
    \scriptsize
    \setlength{\tabcolsep}{3pt}
    \resizebox{\linewidth}{!}{%
    \begin{tabular}{@{}l p{4.2cm} cc cc@{}}
    \toprule
    Model metric & Description & $r_{\mathrm{AR}}$ & $p_{\mathrm{AR}}$ & $r_{\mathrm{score}}$ & $p_{\mathrm{score}}$ \\
    \midrule
    Instruction tuned & Instruction-tuned checkpoint & -0.49 & 0.011 & 0.36 & 0.068  \\
    %Release year & Model release year & -0.48 & 0.014 & 0.13 & 0.516  \\
    Heads & Attention head count & -0.42 & 0.035 & -0.25 & 0.21  \\
    Hidden size & Internal representation width & -0.34 & 0.09 & -0.05 & 0.796  \\
    RMSNorm & RMS-based normalization & -0.30 & 0.141 & 0.24 & 0.244  \\
    GQA & Grouped-query attention & -0.27 & 0.18 & 0.17 & 0.405  \\
    Params & Total model size & -0.27 & 0.183 & 0.10 & 0.641  \\
    Context & Maximum context length & -0.24 & 0.234 & 0.34 & 0.088  \\
    SiLU & Smooth nonlinear activation & -0.20 & 0.325 & -0.27 & 0.176  \\
    SwiGLU & Gated feed-forward activation & -0.18 & 0.378 & 0.36 & 0.071  \\
    Chat tuned & Chat-tuned checkpoint indicator & -0.17 & 0.419 & -0.34 & 0.094  \\
    RoPE & Rotary position encoding & -0.10 & 0.632 & 0.36 & 0.072  \\
    Layers & Transformer layer count & -0.08 & 0.685 & -0.27 & 0.188  \\
    Vocab & Tokenizer vocabulary size & 0.08 & 0.705 & 0.67 & 2.1e-04  \\
    \bottomrule
    \end{tabular}
    }\vspace{-3mm}
    \end{table}

%% file: tables/table_judge_robustness.tex
\begin{table}[!t]
    \centering
    \caption{Judge robustness summary (GPT vs.\ Llama) for HarmBench and Simple Safety Tests. Ambiguity rates (AR), agreement metrics, contradiction types, and rank--range under ambiguity mappings.}
    \label{tab:judge_checks_summary}
    \setlength{\tabcolsep}{3.2pt}
    {\renewcommand{\arraystretch}{1.15}%
    \providecommand{\judgeRotHead}[1]{{\fontsize{5.9}{6.0}\selectfont\rotatebox{40}{\strut #1}}}%
    \begin{tabular}{l c c c c c c c c c}
    \toprule
    \makecell[c]{Suite} & \makecell[c]{AR\\(GPT)} & \makecell[c]{AR\\ (Llama)} & \makecell[c]{$\Delta$AR} & \makecell[c]{Exact\\agree} & \makecell[c]{$\kappa_w$} & \makecell[c]{$0.5\!\leftrightarrow\!$\\$\{0,1\}$} & \makecell[c]{$0\!\leftrightarrow\!1$} & \makecell[c]{Rank-\\range\\(GPT)} & \makecell[c]{Rank-\\range\\(Llama)} \\
    \midrule
    HarmBench & 0.55 & 0.47 & 0.08 & 0.75 & 0.56 & 0.21 & 0.04 & 7.44 & 4.50 \\
    Simple Safety & 0.06 & 0.03 & 0.04 & 0.81 & 0.70 & 0.08 & 0.11 & 0.75 & 1.19 \\
    \bottomrule
    \end{tabular}
    }\vspace{-3mm}
    \end{table}

% \begin{table}[!t]
%     \centering
%     \caption{Judge robustness summary (GPT vs.\ Llama) for HarmBench and Simple Safety Tests on the 16-model overlap where both judge annotations are available. Ambiguity rates (AR), agreement metrics, contradiction types, and rank--range under ambiguity mappings.}
%     \label{tab:judge_checks_summary}
%     \setlength{\tabcolsep}{2pt}
%     {\renewcommand{\arraystretch}{1.15}%
%     \providecommand{\judgeRotHead}[1]{{\fontsize{5.9}{6.0}\selectfont\rotatebox{40}{\strut #1}}}%
%     \begin{tabular}{l r r r r r r r r r}
%     \toprule
%     Suite\rule{0pt}{2.2em} & \judgeRotHead{AR (GPT)} & \judgeRotHead{AR (Llama)} & \judgeRotHead{$\Delta$AR} & \judgeRotHead{Exact agree} & \judgeRotHead{$\kappa_w$} & \judgeRotHead{$0.5\!\leftrightarrow\!\{0,1\}$} & \judgeRotHead{$0\!\leftrightarrow\!1$} & \judgeRotHead{\shortstack{Rank-range\\(GPT)}} & \judgeRotHead{\shortstack{Rank-range\\(Llama)}} \\
%     \midrule
%     HarmBench & 0.55 & 0.47 & 0.08 & 0.75 & 0.56 & 0.21 & 0.04 & 7.44 & 4.50 \\
%     Simple Safety & 0.06 & 0.03 & 0.04 & 0.81 & 0.70 & 0.08 & 0.11 & 0.75 & 1.19 \\
%     \bottomrule
%     \end{tabular}
%     }%
%     \end{table}

%% file: tables/table_judge_confusion_matrices.tex
\begin{table}[!h]
    \centering
    \caption{Confusion matrices between GPT and Llama judges. Cells report count (percent of suite instances). The main pattern is that disagreement concentrates around the ambiguous boundary ($0.5\leftrightarrow\{0,1\}$), especially on HarmBench, rather than direct $0\leftrightarrow 1$ flips. Simple Safety shows stronger diagonal concentration overall, indicating higher cross-judge stability on that suite.}
    \label{tab:judge_confusion_matrices}
    \scriptsize
    \setlength{\tabcolsep}{3.2pt}
    {\renewcommand{\arraystretch}{1.15}%
    \begin{tabular}{l c c c c c c}
    \toprule
     & \multicolumn{3}{c}{HarmBench ($N=6400$)} & \multicolumn{3}{c}{Simple Safety Tests ($N=1600$)} \\
    \cmidrule(lr){2-4}\cmidrule(lr){5-7}
     & Llama 0 & Llama 0.5 & Llama 1 & Llama 0 & Llama 0.5 & Llama 1 \\
    \midrule
    GPT 0   & 234 (3.66\%) & 196 (3.06\%) & 2 (0.03\%) & 640 (40.00\%) & 17 (1.06\%) & 92 (5.75\%) \\
    GPT 0.5 & 502 (7.84\%) & 2594 (40.53\%) & 437 (6.83\%) & 45 (2.81\%) & 10 (0.62\%) & 47 (2.94\%) \\
    GPT 1   & 261 (4.08\%) & 233 (3.64\%) & 1941 (30.33\%) & 85 (5.31\%) & 13 (0.81\%) & 651 (40.69\%) \\
    \bottomrule
    \end{tabular}\vspace{-3mm}
    }%
    \end{table}

%% file: tables/slm_outcome_summary_adjrank.tex
\begin{table}[!h]
    \centering
    \caption{Per-model outcome summary for AirBench, SALAD, and HarmBench. We report the raw ternary-score mean $S_{\mathrm{M}}$, harmful-completion rate (HCR; score $=0$), ambiguity rate (AR; score $=0.5$), safe-refusal rate (SRR; score $=1$), and the ambiguity-adjusted score $S_{adj}=S_{\mathrm{M}}\cdot(1-\mathrm{AR})$. We also report the implied rank change ($\Delta$R) relative to ranking by $S_{\mathrm{M}}$. {\footnotesize In $\Delta$R, $\uparrow$/$\downarrow$ indicates direction, and the number is the absolute number of rank positions moved.}}
    \label{tab:outcome_summary_adjrank}
    \tiny
    \setlength{\tabcolsep}{1pt}
    \resizebox{\linewidth}{!}{%
    \begin{tabular}{l r r r r r r | r r r r r r | r r r r r r}
    \toprule
 & \multicolumn{6}{c}{AirBench} & \multicolumn{6}{c}{SALAD} & \multicolumn{6}{c}{HarmBench} \\
\cmidrule(lr){2-7} \cmidrule(lr){8-13} \cmidrule(lr){14-19}
Model & $S_{\mathrm{M}}$ & HCR & AR & SRR & $S_{adj}$ & $\Delta$R & $S_{\mathrm{M}}$ & HCR & AR & SRR & $S_{adj}$ & $\Delta$R & $S_{\mathrm{M}}$ & HCR & AR & SRR & $S_{adj}$ & $\Delta$R \\
    \midrule
GPT-2 Small & 0.59 & 0.07 & 0.69 & 0.24 & 0.18 & $\downarrow$14 & 0.58 & 0.22 & 0.39 & 0.38 & 0.35 & $\downarrow$2 & 0.56 & 0.01 & 0.86 & 0.13 & 0.08 & $\downarrow$11 \\
OPT-125M & 0.49 & 0.11 & 0.81 & 0.08 & 0.09 & $\downarrow$16 & 0.61 & 0.13 & 0.52 & 0.35 & 0.29 & $\downarrow$15 & 0.58 & 0.01 & 0.82 & 0.17 & 0.11 & $\downarrow$10 \\
Smollm2 135m IT & 0.33 & 0.46 & 0.43 & 0.11 & 0.19 & $\uparrow$2 & 0.46 & 0.47 & 0.14 & 0.39 & 0.39 & $\uparrow$11 & 0.51 & 0.12 & 0.74 & 0.14 & 0.13 & $\uparrow$2 \\
Gemma 3-270M IT & 0.48 & 0.19 & 0.66 & 0.16 & 0.17 & $\downarrow$11 & 0.71 & 0.21 & 0.15 & 0.64 & 0.61 & $\uparrow$2 & 0.68 & 0.01 & 0.61 & 0.38 & 0.26 & $\downarrow$1 \\
OPT-350M & 0.48 & 0.15 & 0.74 & 0.11 & 0.12 & $\downarrow$10 & 0.58 & 0.20 & 0.43 & 0.36 & 0.33 & $\downarrow$5 & 0.57 & 0.01 & 0.84 & 0.15 & 0.09 & $\downarrow$11 \\
GPT-2 Medium & 0.54 & 0.14 & 0.64 & 0.22 & 0.19 & $\downarrow$9 & 0.59 & 0.21 & 0.39 & 0.39 & 0.36 & $\downarrow$3 & 0.55 & 0.03 & 0.84 & 0.13 & 0.09 & $\downarrow$7 \\
Smollm2 360m IT & 0.27 & 0.62 & 0.23 & 0.15 & 0.21 & $\uparrow$8 & 0.54 & 0.41 & 0.11 & 0.48 & 0.47 & $\uparrow$10 & 0.42 & 0.30 & 0.55 & 0.14 & 0.19 & $\uparrow$8 \\
Qwen 2.5-0.5B IT & 0.37 & 0.57 & 0.12 & 0.31 & 0.33 & $\uparrow$10 & 0.67 & 0.23 & 0.21 & 0.56 & 0.52 & $\uparrow$1 & 0.63 & 0.22 & 0.30 & 0.48 & 0.44 & $\uparrow$1 \\
GPT-2 Large & 0.46 & 0.23 & 0.62 & 0.15 & 0.18 & $\downarrow$4 & 0.56 & 0.28 & 0.32 & 0.40 & 0.38 & $\uparrow$4 & 0.56 & 0.09 & 0.70 & 0.21 & 0.17 & $\downarrow$4 \\
TinyLlama 1.1B chat & 0.19 & 0.73 & 0.16 & 0.11 & 0.16 & $\uparrow$3 & 0.59 & 0.27 & 0.27 & 0.46 & 0.43 & $\uparrow$2 & 0.38 & 0.38 & 0.49 & 0.14 & 0.19 & $\uparrow$11 \\
OLMo 2 0425 1B & 0.60 & 0.15 & 0.49 & 0.36 & 0.31 & $\downarrow$6 & 0.50 & 0.31 & 0.37 & 0.32 & 0.32 & $\uparrow$2 & 0.55 & 0.04 & 0.82 & 0.14 & 0.10 & $\downarrow$3 \\
MobileLLM-1B & 0.47 & 0.15 & 0.75 & 0.10 & 0.12 & $\downarrow$10 & 0.46 & 0.37 & 0.33 & 0.29 & 0.31 & $\uparrow$1 & 0.55 & 0.02 & 0.85 & 0.12 & 0.08 & $\downarrow$7 \\
Gemma 3-1B IT & 0.51 & 0.30 & 0.39 & 0.32 & 0.31 & -- & 0.75 & 0.03 & 0.44 & 0.53 & 0.42 & $\downarrow$10 & 0.79 & 0.01 & 0.40 & 0.59 & 0.47 & $\downarrow$4 \\
Llama 3.2-1B IT & 0.54 & 0.42 & 0.08 & 0.50 & 0.49 & $\uparrow$3 & 0.71 & 0.20 & 0.19 & 0.62 & 0.58 & $\uparrow$2 & 0.79 & 0.10 & 0.23 & 0.68 & 0.61 & $\uparrow$1 \\
Zamba2 1.2B IT & 0.48 & 0.47 & 0.11 & 0.42 & 0.43 & $\uparrow$6 & 0.61 & 0.29 & 0.21 & 0.50 & 0.48 & $\uparrow$3 & 0.72 & 0.12 & 0.32 & 0.56 & 0.49 & -- \\
DeepSeek-R1-Qwen & 0.25 & 0.62 & 0.26 & 0.12 & 0.19 & $\uparrow$5 & 0.54 & 0.28 & 0.37 & 0.35 & 0.34 & $\uparrow$2 & 0.51 & 0.03 & 0.92 & 0.05 & 0.04 & $\downarrow$7 \\
GPT-2 XL & 0.46 & 0.25 & 0.58 & 0.17 & 0.19 & $\uparrow$1 & 0.57 & 0.27 & 0.33 & 0.40 & 0.38 & $\uparrow$2 & 0.56 & 0.09 & 0.69 & 0.22 & 0.18 & $\downarrow$3 \\
Qwen 2.5-1.5B IT & 0.61 & 0.36 & 0.07 & 0.57 & 0.56 & $\uparrow$1 & 0.88 & 0.04 & 0.16 & 0.81 & 0.75 & -- & 0.82 & 0.13 & 0.10 & 0.77 & 0.74 & $\uparrow$1 \\
Stablelm 2 1 6b & 0.70 & 0.12 & 0.37 & 0.51 & 0.44 & $\downarrow$5 & 0.64 & 0.25 & 0.22 & 0.53 & 0.50 & $\uparrow$1 & 0.56 & 0.05 & 0.79 & 0.17 & 0.12 & $\downarrow$4 \\
Smollm2 1.7b IT & 0.32 & 0.62 & 0.11 & 0.27 & 0.29 & $\uparrow$9 & 0.63 & 0.29 & 0.14 & 0.56 & 0.54 & $\uparrow$4 & 0.42 & 0.36 & 0.44 & 0.20 & 0.24 & $\uparrow$11 \\
danube2 1.8b chat & 0.25 & 0.70 & 0.10 & 0.20 & 0.23 & $\uparrow$11 & 0.58 & 0.33 & 0.18 & 0.49 & 0.48 & $\uparrow$8 & 0.32 & 0.51 & 0.34 & 0.14 & 0.21 & $\uparrow$14 \\
Gemma 2-2B IT & 0.67 & 0.24 & 0.17 & 0.59 & 0.56 & $\uparrow$1 & 0.69 & 0.04 & 0.54 & 0.42 & 0.32 & $\downarrow$17 & 0.86 & 0.01 & 0.26 & 0.73 & 0.64 & $\downarrow$1 \\
Llama 3.2-3B IT & 0.53 & 0.43 & 0.07 & 0.50 & 0.49 & $\uparrow$5 & 0.51 & 0.28 & 0.41 & 0.31 & 0.30 & $\downarrow$2 & 0.69 & 0.22 & 0.18 & 0.59 & 0.56 & $\uparrow$2 \\
Qwen 2.5-3B IT & 0.48 & 0.47 & 0.08 & 0.44 & 0.44 & $\uparrow$7 & 0.77 & 0.09 & 0.29 & 0.62 & 0.55 & $\downarrow$2 & 0.78 & 0.12 & 0.22 & 0.67 & 0.61 & $\uparrow$1 \\
Dolly v2 3b & 0.23 & 0.63 & 0.27 & 0.10 & 0.17 & $\uparrow$4 & 0.61 & 0.24 & 0.31 & 0.45 & 0.42 & $\downarrow$4 & 0.40 & 0.30 & 0.61 & 0.09 & 0.16 & $\uparrow$6 \\
danube3.1 4b chat & 0.28 & 0.66 & 0.11 & 0.23 & 0.25 & $\uparrow$9 & 0.60 & 0.30 & 0.20 & 0.50 & 0.48 & $\uparrow$5 & 0.36 & 0.47 & 0.35 & 0.18 & 0.23 & $\uparrow$14 \\
    \bottomrule
    \end{tabular}%
    }
\end{table}

%% file: tables/table_models_compact_copy.tex
\begin{table}[h]
    \centering
    \caption{SLMs evaluated in this study (26 models spanning 124M--4.0B parameters) and the model metadata. %used in our stratified and correlational analyses. %Reported architecture and training-data summaries are taken from model cards and technical reports (``--'' indicates not publicly specified). 
    The table highlights substantial heterogeneity in size, context length, tuning, and architectural choices, motivating family- and metadata-aware interpretation of benchmark outcomes.}
    \label{tab:models}
    \small
    \setlength{\tabcolsep}{1.5pt}
    \resizebox{\linewidth}{!}{%
    \begin{tabular}{@{}p{3.4cm}|p{1.0cm}|p{7.30cm}|p{5.0cm} |p{2.5cm}@{}}
    \toprule
    Vendor, year / Model & Params / Ctx & Architecture (reported) & Training data (reported) & {\parbox[t]{\linewidth}{\raggedright{\raggedright Objective (reported)}}} \\
    \midrule
    {\parbox[t]{\linewidth}{\raggedright OpenAI 2019\\\mbox{GPT-2 Small~\cite{hfGpt2Small}}}} & {\parbox[t]{\linewidth}{\centering 124M\\1024}} & Decoder-only transformer; 12 layers; 12 attention heads; 768 hidden; 3072 FFN; GELU; absolute position embeddings; vocab 50257 & WebText $\sim$40GB from 8M web pages; cutoff Dec 2017 & Causal language modeling \\\hline
    {\parbox[t]{\linewidth}{\raggedright Meta 2022\\\mbox{OPT-125M~\cite{hfOpt125m}}}} & {\parbox[t]{\linewidth}{\centering 125M\\2048}} & Decoder-only transformer (OPT/GPT-style); 12 layers; 12 heads; 768 hidden; 3072 FFN; ReLU; max pos 2048; vocab 50272 & 180B tokens; BookCorpus; Common Crawl; Reddit; predominantly English & Causal language modeling \\\hline
    {\parbox[t]{\linewidth}{\raggedright HF 2024\\\mbox{SmolLM2-135M IT~\cite{hfSmolLM2135mInstruct}}}} & {\parbox[t]{\linewidth}{\centering 135M\\8192}} & Transformer decoder (Llama-family); 30 layers; 9 heads; 3 KV heads; 576 hidden; 1536 FFN; vocab 49152 & 2T-token curated pre-training mix; instruction-tuning data & Pre-training + SFT + DPO \\\hline
    {\parbox[t]{\linewidth}{\raggedright Google 2025\\\mbox{Gemma3-270M IT~\cite{hfGemma3270mIt}}}} & {\parbox[t]{\linewidth}{\centering 270M\\32768}} & Decoder-only; 270M total (170M embedding from 256K vocab + 100M transformer blocks) &  2T tokens; web; code; math; 140+ languages & Pre-training + instruction tuning \\\hline
    {\parbox[t]{\linewidth}{\raggedright Meta 2022\\\mbox{OPT-350M~\cite{hfOpt350m}}}} & {\parbox[t]{\linewidth}{\centering 350M\\2048}} & Decoder-only transformer (OPT/GPT-style); 24 layers; 16 heads; 1024 hidden; 4096 FFN; ReLU; max pos 2048; vocab 50272 & 180B tokens; BookCorpus; Common Crawl; Reddit; predominantly English & Causal language modeling \\\hline
    {\parbox[t]{\linewidth}{\raggedright OpenAI 2019\\\mbox{GPT-2 Medium~\cite{hfGpt2Medium}}}} & {\parbox[t]{\linewidth}{\centering 355M\\1024}} & Decoder-only transformer; 24 layers; 16 attention heads; 1024 hidden; 4096 FFN; GELU; absolute position embeddings; vocab 50257 &  WebText $\sim$40GB from 8M web pages; cutoff Dec 2017 & Causal language modeling \\\hline
    {\parbox[t]{\linewidth}{\raggedright HF 2024\\\mbox{SmolLM2-360M IT~\cite{hfSmolLM2360mInstruct}}}} & {\parbox[t]{\linewidth}{\centering 360M\\8192}} & Transformer decoder (Llama-family); 32 layers; 15 heads; 5 KV heads; 960 hidden; 2560 FFN; vocab 49152 & 4T-token curated pre-training mix; instruction-tuning data & Pre-training + SFT + DPO \\\hline
    {\parbox[t]{\linewidth}{\raggedright Alibaba 2024\\\mbox{Qwen2.5-0.5B IT~\cite{hfQwen25Zero5bInstruct}}}} & {\parbox[t]{\linewidth}{\centering 500M\\32768}} & Decoder-only; 24 layers; 896 hidden; 14 heads 2 KV (GQA); RoPE; SwiGLU; RMSNorm; vocab 151936 & 18T tokens; web; code; math; synthetic; 29+ languages; 1M+ SFT & Pre-training + DPO/GRPO \\\hline
    {\parbox[t]{\linewidth}{\raggedright OpenAI 2019\\\mbox{GPT-2 Large~\cite{hfGpt2Large}}}} & {\parbox[t]{\linewidth}{\centering 774M\\1024}} & Decoder-only transformer; 36 layers; 20 attention heads; 1280 hidden; 5120 FFN; GELU; absolute position embeddings; vocab 50257 & WebText $\sim$40GB from 8M web pages; cutoff Dec 2017 & Causal language modeling \\\hline
    {\parbox[t]{\linewidth}{\raggedright Google 2025\\\mbox{Gemma3-1B IT~\cite{hfGemma31bIt}}}} & {\parbox[t]{\linewidth}{\centering 1.0B\\32768}} & Decoder-only; $\sim$1.1B active params; 32K context; text-only input & 2T tokens; web; code; math; 140+ languages; cutoff Aug 2024 & Pre-training + instruction tuning \\\hline
    {\parbox[t]{\linewidth}{\raggedright Meta 2024\\\mbox{MobileLLM-1B~\cite{hfMobileLLM1b}}}} & {\parbox[t]{\linewidth}{\centering 1.0B\\--}} & Deep thin decoder-only; embedding sharing; grouped-query attention; optional block-wise weight-sharing; sub-billion optimized & Training data not specified in paper; architecture-focused design & Pre-training + chat tuning \\\hline
    {\parbox[t]{\linewidth}{\raggedright AI2 2025\\\mbox{OLMo2-0425-1B~\cite{hfOlmo204251b}}}} & {\parbox[t]{\linewidth}{\centering 1.0B\\4096}} & Transformer decoder (OLMo2); 16 layers; 16 heads; 2048 hidden; 8192 FFN; RoPE; vocab 100352 & OLMo-mix-1124 with Dolmino-mix-1124 mid-training & Causal language modeling \\\hline
    {\parbox[t]{\linewidth}{\raggedright TinyLlama. 2023\\\mbox{1.1B Chat v1.0~\cite{hfTinyLlama11bChat}}}} & {\parbox[t]{\linewidth}{\centering 1.1B\\2048}} & Llama-style decoder-only; 22 layers; 32 heads; 4 KV heads (GQA); 2048 hidden; 5632 FFN; RoPE; RMSNorm; vocab 32000 & SlimPajama + StarCoderData; aligned with UltraChat/UltraFeedback & Causal pre-training + SFT + DPO \\\hline
    {\parbox[t]{\linewidth}{\raggedright Zyphra 2024\\\mbox{Zamba2-1.2B IT~\cite{hfZamba212bInstruct}}}} & {\parbox[t]{\linewidth}{\centering 1.2B\\4096}} & Hybrid SSM/transformer (Mamba2 + shared attention blocks); 38 layers; 32 attention heads; 2048 hidden; vocab 32000 & UltraChat-200k, Infinity-Instruct, UltraFeedback, preference data & SFT + DPO instruction tuning \\\hline
    {\parbox[t]{\linewidth}{\raggedright Meta 2024\\\mbox{Llama3.2-1B IT~\cite{hfLlama321bInstruct}}}} & {\parbox[t]{\linewidth}{\centering 1.23B\\131072}} & Auto-regressive decoder-only transformer; GQA; shared embeddings; 128K context; 1.23B params & Public online data; up to 9T tokens pretraining; cutoff Dec 2023 & SFT + RLHF \\\hline
    {\parbox[t]{\linewidth}{\raggedright OpenAI 2019\\\mbox{GPT-2 XL~\cite{hfGpt2XL}}}} & {\parbox[t]{\linewidth}{\centering 1.5B\\1024}} & Decoder-only transformer; 48 layers; 25 attention heads; 1600 hidden; 6400 FFN; GELU; absolute position embeddings; vocab 50257 & WebText $\sim$40GB from 8M web pages; cutoff Dec 2017 & Causal language modeling \\\hline
    {\parbox[t]{\linewidth}{\raggedright DeepSeek 2024\\\mbox{R1-Qwen-1.5B~\cite{hfDeepSeekR1DistillQwen15b}}}} & {\parbox[t]{\linewidth}{\centering 1.5B\\131072}} & Qwen2.5-Math-1.5B base: 28 layers; 1536 hidden; 12 heads 2 KV (GQA); RoPE; SiLU; RMSNorm; vocab 151936; 131K context & $\sim$800K SFT samples; distilled from DeepSeek-R1 reasoning model & Distillation from DeepSeek-R1 (reasoning/CoT) \\\hline
    {\parbox[t]{\linewidth}{\raggedright Alibaba 2024\\\mbox{Qwen2.5-1.5B IT~\cite{hfQwen2515bInstruct}}}} & {\parbox[t]{\linewidth}{\centering 1.5B\\131072}} & Decoder-only; 28 layers; 1536 hidden; 12 heads 2 KV (GQA); RoPE; SwiGLU; RMSNorm; vocab 151936 & 18T tokens; web; code; math; synthetic; 29+ languages; 1M+ SFT & Pre-training + DPO/GRPO \\\hline
    {\parbox[t]{\linewidth}{\raggedright StabAI 2024\\\mbox{StableLM2-1.6B~\cite{hfStablelm216b}}}} & {\parbox[t]{\linewidth}{\centering 1.6B\\4096}} & StableLM decoder-only; 24 layers; 32 heads; 2048 hidden; 5632 FFN; partial RoPE; vocab 100352 & 2T-token multilingual and code-heavy pre-training mix & Casual language modeling\\\hline
    {\parbox[t]{\linewidth}{\raggedright HF 2024\\\mbox{SmolLM2-1.7B IT~\cite{hfSmolLM217bInstruct}}}} & {\parbox[t]{\linewidth}{\centering 1.7B\\8192}} & Transformer decoder (Llama-family); 24 layers; 32 heads; 2048 hidden; vocab 49152 & 11T-token mix: FineWeb-Edu, DCLM, The Stack, math/code & Pre-training + SFT + DPO \\\hline
    {\parbox[t]{\linewidth}{\raggedright H2O 2024\\\mbox{Danube2-1.8B Chat~\cite{hfDanube218bChat}}}} & {\parbox[t]{\linewidth}{\centering 1.8B\\8192}} & Llama2/Mistral-style decoder-only; 24 layers; 32 heads; 8 KV heads (GQA); 2560 hidden; SiLU; RMSNorm; vocab 32000 & H2O Danube2 pre-training corpus; H2O LLM Studio chat data & Pre-training + SFT + DPO \\\hline
    {\parbox[t]{\linewidth}{\raggedright Google 2024\\\mbox{Gemma2-2B IT~\cite{hfGemma22bIt}}}} & {\parbox[t]{\linewidth}{\centering 2.2B\\8192}} & Decoder-only; 26 layers; 2048 hidden; 16 attention heads; 4 key-value heads; GQA; RoPE; RMSNorm; GELU & 2T tokens; 8K context during pre-training; knowledge distillation & Knowledge distillation + pre-training \\\hline
    {\parbox[t]{\linewidth}{\raggedright Alibaba 2024\\\mbox{Qwen2.5-3B IT~\cite{hfQwen253bInstruct}}}} & {\parbox[t]{\linewidth}{\centering 3.0B\\32768}} & Decoder-only; 36 layers; 2048 hidden; 16 heads 2 KV (GQA); RoPE; SwiGLU; RMSNorm; vocab 151936 & 18T tokens; web; code; math; synthetic; 29+ languages; 1M+ SFT & Pre-training + DPO/GRPO \\\hline
    {\parbox[t]{\linewidth}{\raggedright Dbricks 2023\\\mbox{Dolly-v2-3B~\cite{hfDollyV23b}}}} & {\parbox[t]{\linewidth}{\centering 3.0B\\2048}} & GPT-NeoX/Pythia-style decoder-only; derived from EleutherAI Pythia-2.8B & Databricks-dolly-15k instruction dataset & Instruction tuning on Pythiabase \\\hline
    {\parbox[t]{\linewidth}{\raggedright Meta 2024\\\mbox{Llama3.2-3B IT~\cite{hfLlama323bInstruct}}}} & {\parbox[t]{\linewidth}{\centering 3.21B\\131072}} & Auto-regressive decoder-only transformer; GQA; shared embeddings & Public online data; up to 9T tokens pretraining; cutoff Dec 2023 & SFT + RLHF \\\hline
    {\parbox[t]{\linewidth}{\raggedright H2O 2024\\\mbox{Danube3.1-4B Chat~\cite{hfDanube314bChat}}}} & {\parbox[t]{\linewidth}{\centering 4.0B\\8192}} & Llama-style decoder-only; 24 layers; 32 heads; 8 KV heads (GQA); 3840 hidden; SiLU; RMSNorm & H2O Danube3 pre-training corpus; H2O LLM Studio chat data & Pre-training + chat fine-tuning \\\hline
    \bottomrule
    \end{tabular}%
    }
    \end{table}

%% file: reference.bib
@article{subramanian2025slm,
  author = {Subramanian, S. and Elango, V. and Gungor, M.},
  title = {Small Language Models (SLMs) Can Still Pack a Punch: A Survey},
  journal = {arXiv preprint arXiv:2501.05465},
  year = {2025}
}

@article{ren2024safetywashing,
  _author = {Ren, Richard and Basart, Steven and Khoja, Adam and Pan, Alexander and Gatti, Alice and Phan, Long and Yin, Xuwang and Mazeika, Mantas and Mukobi, Gabriel and Kim, Ryan Hwang and Fitz, Stephen and Hendrycks, Dan},
  author = {Ren, Richard and Basart, Steven and Khoja, Adam and Pan, Alexander and Gatti, Alice and others},
  title = {Safetywashing: Do AI Safety Benchmarks Actually Measure Safety Progress?},
  journal={NeurIPS},
  year = {2024}
}

@inproceedings{10.1145/3711896.3736563,
  author = {Wang, Fali and Lin, Minhua and Ma, Yao and Liu, Hui and He, Qi and Tang, Xianfeng and Tang, Jiliang and Pei, Jian and Wang, Suhang},
  _author = {Wang, Fali and Lin, Minhua and Ma, Yao and Liu, Hui and He, Qi and Tang, Xianfeng and Tang, Jiliang and others},
  title = {A Survey on Small Language Models in the Era of Large Language Models: Architecture, Capabilities, and Trustworthiness},
  booktitle = {ACM SIGKDD},
  year = {2025}
}

@article{garg2025rise,
  author = {Garg, Muskan and Raza, Shaina and Rayana, Shebuti and Liu, Xingyi and Sohn, Sunghwan},
  title = {The Rise of Small Language Models in Healthcare: A Comprehensive Survey},
  journal = {arXiv preprint arXiv:2504.17119},
  year = {2025}
}

@article{slm_survey,
  _author = {Nguyen, Chien Van and Shen, Xuan and Aponte, Ryan and Xia, Yu and Basu, Samyadeep and Hu, Zhengmian and Chen, Jian and Parmar, Mihir and Kunapuli, Sasidhar and Barrow, Joe and Wu, Junda and Singh, Ashish and Wang, Yu and Gu, Jiuxiang and Dernoncourt, Franck and Ahmed, Nesreen K. and Lipka, Nedim and Zhang, Ruiyi and Chen, Xiang and Yu, Tong and Kim, Sungchul and Deilamsalehy, Hanieh and Park, Namyong and Rimer, Mike and Zhang, Zhehao and Yang, Huanrui and Rossi, Ryan A. and Nguyen, Thien Huu},
  author = {Nguyen, Chien Van and Shen, Xuan and Aponte, Ryan and Xia, Yu and Basu, Samyadeep and Hu, Zhengmian and others},
  title = {A Survey of Small Language Models},
  journal = {arXiv preprint arXiv:2410.20011},
  year = {2024}
}

@inproceedings{li2024salad,
  _author = {Li, Lijun and Dong, Bowen and Wang, Ruohui and Hu, Xuhao and Zuo, Wangmeng and others},
  author = {Li, Lijun and Dong, Bowen and Wang, Ruohui and Hu, Xuhao and Zuo, Wangmeng and Lin, Dahua and Qiao, Yu and Shao, Jing},
  title = {Salad-bench: A hierarchical and comprehensive safety benchmark for large language models},
  booktitle = {Findings of the Association for Computational Linguistics: ACL 2024},
  year = {2024}
}

@inproceedings{parrish-etal-2022-bbq,
  title={{BBQ}: A hand-built bias benchmark for question answering},
  author={Parrish, Alicia and Chen, Angelica and Nangia, Nikita and Padmakumar, Vishakh and Phang, Jason and Thompson, Jana and Htut, Phu Mon and Bowman, Samuel},
  _author = {Parrish, Alicia and Chen, Angelica and Nangia, Nikita and Padmakumar, Vishakh and Phang, Jason and Thompson, Jana and others},
  booktitle={Findings of ACL},
  year={2022}
}

@inproceedings{mazeika2024harmbench,
  author = {Mazeika, Mantas and Phan, Long and Yin, Xuwang and Zou, Andy and Wang, Zifan and Mu, Norman and Sakhaee, Elham and Li, Nathaniel and Basart, Steven and Li, Bo and others},
  _author = {Mazeika, Mantas and Phan, Long and Yin, Xuwang and Zou, Andy and Wang, Zifan and Mu, Norman and others},
  title = {HarmBench: A Standardized Evaluation Framework for Automated Red Teaming and Robust Refusal},
  booktitle = {ICML},
  year = {2024}
}

@article{overview_slm,
author = {Wang, Fali and Zhang, Zhiwei and others},
_author = {Wang, Fali and Zhang, Zhiwei and Zhang, Xianren and Wu, Zongyu and Mo, TzuHao and Lu, Qiuhao and Wang, Wanjing and Li, Rui and Xu, Junjie and Tang, Xianfeng and He, Qi and Ma, Yao and Huang, Ming and Wang, Suhang},
title = {A Comprehensive Survey of Small Language Models in the Era of Large Language Models: Techniques, Enhancements, Applications, Collaboration with LLMs, and Trustworthiness},
journal = {ACM Trans. Intell. Syst. Technol.},
year = {2025}
}

@article{role_slm,
  author = {Lihu Chen and Gaël Varoquaux},
  title = {What is the Role of Small Models in the LLM Era: A Survey},
  journal = {arXiv preprint arXiv:2409.06857},
  year = {2025}
}

@article{chen2025survey,
  author = {Chen, Yi and Zhao, JiaHao and Han, HaoHao},
  title = {A survey on collaborative mechanisms between large and small language models},
  journal = {arXiv preprint arXiv:2505.07460},
  year = {2025}
}

@article{ai_benchmarks,
  author = {Ivanov, Todor and Penchev, Valeri},
  title = {AI Benchmarks and Datasets for LLM Evaluation},
  journal = {arXiv preprint arXiv:2412.01020},
  year = {2024}
}

@inproceedings{li2024privlm,
  _author = {Li, Haoran and Guo, Dadi and Li, Donghao and Fan, Wei and Hu, Qi and Liu, Xin and Chan, Chunkit and Yao, Duanyi and Yao, Yuan and Song, Yangqiu},
  author = {Li, Haoran and Guo, Dadi and Li, Donghao and Fan, Wei and Hu, Qi and Liu, Xin and Chan, Chunkit and others},
  title = {Privlm-bench: A multi-level privacy evaluation benchmark for language models},
  booktitle = {ACL},
  year = {2024}
}

@article{airbench2024,
  _author = {Zeng, Yi and Yang, Yu and Zhou, Andy and Tan, Jeffrey Ziwei and Tu, Yuheng and Mai, Yifan and Klyman, Kevin and Pan, Minzhou and Jia, Ruoxi and Song, Dawn and others},
  author = {Zeng, Yi and Yang, Yu and Zhou, Andy and Tan, Jeffrey Ziwei and Tu, Yuheng and Mai, Yifan and Klyman, Kevin and others},
  title = {Air-bench 2024: A safety benchmark based on risk categories from regulations and policies},
  journal={AGI-Artificial General Intelligence-Robotics-Safety \& Alignment},
  year = {2024}
}

@article{simplesafetytests,
  author = {Vidgen, Bertie and Scherrer, Nino and Kirk, Hannah Rose and Qian, Rebecca and Kannappan, Anand and Hale, Scott A. and Röttger, Paul},
  _author = {Vidgen, Bertie and Scherrer, Nino and Kirk, Hannah Rose and Qian, Rebecca and others},
  title = {SimpleSafetyTests: A Test Suite for Evaluating LLM Safety},
  journal = {arXiv preprint arXiv:2311.08370},
  year = {2023}
}

@article{helmsafety,
  title={Helm safety: Towards standardized safety evaluations of language models},
  author={Kaiyom, Farzaan and Ahmed, Ahmed and Mai, Yifan and Klyman, Kevin and Bommasani, Rishi and Liang, Percy},
  _author={Kaiyom, Farzaan and Ahmed, Ahmed and others},
  journal={Stanford Center for Research on Foundation Models},
  year={2024}
}

@article{ouyang2022,
  author = {Ouyang, Long and Wu, Jeffrey and Jiang, Xu and Almeida, Diogo and Wainwright, Carroll and Mishkin, Pamela and Zhang, Chong and Agarwal, Sandhini and others},
  _author={Ouyang, Long and Wu, Jeffrey and Jiang, Xu and Almeida, Diogo and Wainwright, Carroll and Mishkin, Pamela and Zhang, Chong and Agarwal, Sandhini and Slama, Katarina and Ray, Alex and others},
  title = {Training language models to follow instructions with human feedback},
  journal={NeurIPS},
  year = {2022}
}

@article{brown2020,
  author = {Brown, Tom and Mann, Benjamin and Ryder, Nick and Subbiah, Melanie and others},
  title = {Language models are few-shot learners},
  _author={Brown, Tom and Mann, Benjamin and Ryder, Nick and Subbiah, Melanie and Kaplan, Jared D and Dhariwal, Prafulla and Neelakantan, Arvind and Shyam, Pranav and Sastry, Girish and Askell, Amanda and others},
  journal={NeurIPS},
  year = {2020}
}

@misc{zheng2023,
  title={Judging llm-as-a-judge with mt-bench and chatbot arena},
  author={Zheng, Lianmin and Chiang, Wei-Lin and Sheng, Ying and Zhuang, Siyuan and Wu, Zhanghao and Zhuang, Yonghao and others},
  _author={Zheng, Lianmin and Chiang, Wei-Lin and Sheng, Ying and Zhuang, Siyuan and Wu, Zhanghao and Zhuang, Yonghao and Lin, Zi and Li, Zhuohan and Li, Dacheng and Xing, Eric and others},
  journal={Advances in neural information processing systems},
  year={2023}
}

@inproceedings{dumitrache2018crowdtruth,
  author = {Dumitrache, Anca and Inel, Oana and Aroyo, Lora and Timmermans, Benjamin and Welty, Chris},
  title = {CrowdTruth 2.0: Quality Metrics for Crowdsourcing with Disagreement},
  year = {2018}
}

@inproceedings{leonardelli2021agreeing,
  author = {Leonardelli, Elisa and Basile, Valerio and Poesio, Massimo and Uma{\~n}a, Martina and Stojanov, Drago{\c{s}}},
  _author = {Leonardelli, Elisa and Basile, Valerio and others},
  title = {Agreeing to Disagree: Annotating Offensive Language Datasets with Annotators' Disagreement},
  booktitle = {EMNLP},
  year = {2021}
}

@inproceedings{liu2023geval,
  _author = {Liu, Yang and Iter, Dan and Xu, Yichong and Wang, Shuohang and Xu, Ruochen and Zhu, Chenguang},
  author = {Liu, Yang and Iter, Dan and Xu, Yichong and Wang, Shuohang and others},
  title = {{G}-{E}val: {NLG} Evaluation using {GPT}-4 with Better Human Alignment},
  booktitle = {EMNLP},
  year = {2023}
}

@inproceedings{li2024crowdsourced,
  title   = {From Crowdsourced Data to High-Quality Benchmarks: Arena-Hard and BenchBuilder Pipeline},
  _author  = {Li, Tianle and Chiang, Wei-Lin and Frick, Evan and Dunlap, Lisa and Wu, Tianhao and Zhu, Banghua and others},
  author  = {Li, Tianle and Chiang, Wei-Lin and Frick, Evan and Dunlap, Lisa and Wu, Tianhao and Zhu, Banghua and Gonzalez, Joseph E. and Stoica, Ion},
  booktitle = {ICML},
  year    = {2025}
}

@article{lin2024wildbench,
  title   = {WildBench: Benchmarking LLMs with Challenging Tasks from Real Users in the Wild},
  _author  = {Lin, Bill Yuchen and Deng, Yuntian and Chandu, Khyathi and Brahman, Faeze and Ravichander, Abhilasha and Pyatkin, Valentina and Dziri, Nouha and Le Bras, Ronan and Choi, Yejin},
  author  = {Lin, Bill Yuchen and Deng, Yuntian Chandu, Khyathi and Brahman, Faeze and Ravichander, Abhilasha and Pyatkin, Valentina and Dziri, Nouha and others},
  journal = {arXiv preprint arXiv:2406.04770},
  year    = {2024}
}

@inproceedings{white2024livebench,
  title   = {LiveBench: A Challenging, Contamination-Limited LLM Benchmark},
  _author  = {White, Colin and Dooley, Samuel and Roberts, Manley and Pal, Arka and Feuer, Ben and Jain, Siddhartha and Shwartz-Ziv, Ravid and Jain, Neel and Saifullah, Khalid and Dey, Sreemanti and Agrawal, Shubh and Sandha, Sandeep Singh and Naidu, Siddartha and Hegde, Chinmay and LeCun, Yann and Goldstein, Tom and Neiswanger, Willie and Goldblum, Micah},
  author  = {White, Colin and Dooley, Samuel and Roberts, Manley and Pal, Arka and Feuer, Ben and Jain, Siddhartha and Shwartz-Ziv, Ravid and Jain, Neel and others},
  booktitle = {ICLR},
  year    = {2025}
}

@inproceedings{ni2024mixeval,
  title   = {MixEval: Deriving Wisdom of the Crowd from LLM Benchmark Mixtures},
  author  = {Ni, Jinjie and Xue, Fuzhao and Yue, Xiang and Deng, Yuntian and Shah, Mahir and Jain, Kabir and Neubig, Graham and You, Yang},
  _author  = {Ni, Jinjie and Xue, Fuzhao and others},
  booktitle = {NeurIPS},
  year    = {2024}
}

@article{zhou2023ifeval,
  title   = {Instruction-Following Evaluation for Large Language Models},
  _author  = {Zhou, Jeffrey and Lu, Tianjian and Mishra, Swaroop and Brahma, Siddhartha and Basu, Sujoy and Luan, Yi and Zhou, Denny and Hou, Le},
  author  = {Zhou, Jeffrey and Lu, Tianjian and Mishra, Swaroop and Brahma, Siddhartha and Basu, Sujoy and others},
  journal = {arXiv preprint arXiv:2311.07911},
  year    = {2023}
}

@inproceedings{wei2023jailbroken,
  author = {Wei, Alexander and Haghtalab, Nika and Steinhardt, Jacob},
  title = {Jailbroken: How Does LLM Safety Training Fail?},
  booktitle = {NeurIPS},
  year = {2023}
}

@misc{zou2023universal,
  author = {Zou, Andy and Wang, Zifan and Carlini, Nicholas and Nasr, Milad and Kolter, J. Zico and Fredrikson, Matt},
  title = {Universal and Transferable Adversarial Attacks on Aligned Language Models},
  howpublished = {arXiv preprint arXiv:2307.15043},
  year = {2023}
}

@misc{euAIAct2024,
  author = {{European Parliament and Council of the European Union}},
  title = {Regulation (EU) 2024/1689 Laying Down Harmonised Rules on Artificial Intelligence},
  howpublished = {Official Journal of the European Union},
  year = {2024}
}

@misc{nistAIrmf2023,
  author = {{National Institute of Standards and Technology}},
  title = {Artificial Intelligence Risk Management Framework (AI RMF 1.0)},
  howpublished = {NIST AI 100-1},
  year = {2023}
}

@misc{chinaGenAIMeasures2023,
  author = {{Cyberspace Administration of China}},
  title = {Interim Measures for the Management of Generative Artificial Intelligence Services},
  howpublished = {Cyberspace Administration of China},
  year = {2023}
}

@article{breneman2024readability,
  author = {Breneman, A. and Trager, M. H. and Gordon, E. R. and Samie, F. H.},
  title = {Readability rescue: large language models may improve readability of patient education materials},
  journal = {Archives of Dermatological Research},
  volume = {316},
  number = {9},
  pages = {669},
  year = {2024},
}

@article{marulli2024readability,
  author = {Marulli, Fiammetta and Campanile, Lelio and de Biase, Maria Stella and Marrone, Stefano and Verde, Laura and Bifulco, Marianna},
  title = {Understanding readability of large language models output: an empirical analysis},
  journal = {Procedia Computer Science},
  volume = {246},
  pages = {5273--5282},
  year = {2024},
  _url = {https://www.sciencedirect.com/science/article/pii/S1877050924026905}
}

@inproceedings{reimers2019sentencebert,
  author = {Reimers, Nils and Gurevych, Iryna},
  title = {Sentence-BERT: Sentence Embeddings using Siamese BERT-Networks},
  booktitle = {EMNLP},
  year = {2019}
}

@misc{hfGpt2Small,
  author = {{OpenAI}},
  title = {gpt2 Model Card},
  howpublished = {Hugging Face},
  year = {2019}
}

@misc{hfGpt2Medium,
  author = {{OpenAI}},
  title = {gpt2-medium Model Card},
  howpublished = {Hugging Face},
  year = {2019}
}

@misc{hfGpt2Large,
  author = {{OpenAI}},
  title = {gpt2-large Model Card},
  howpublished = {Hugging Face},
  year = {2019}
}

@misc{hfGpt2XL,
  author = {{OpenAI}},
  title = {gpt2-xl Model Card},
  howpublished = {Hugging Face},
  year = {2019}
}

@misc{hfOpt125m,
  author = {{Meta AI}},
  title = {facebook/opt-125m Model Card},
  howpublished = {Hugging Face},
  year = {2022}
}

@misc{hfOpt350m,
  author = {{Meta AI}},
  title = {facebook/opt-350m Model Card},
  howpublished = {Hugging Face},
  year = {2022}
}

@misc{hfGemma3270mIt,
  author = {{Google}},
  title = {google/gemma-3-270m-it Model Card},
  howpublished = {Hugging Face},
  year = {2025}
}

@misc{hfGemma31bIt,
  author = {{Google}},
  title = {google/gemma-3-1b-it Model Card},
  howpublished = {Hugging Face},
  year = {2025}
}

@misc{hfGemma22bIt,
  author = {{Google}},
  title = {google/gemma-2-2b-it Model Card},
  howpublished = {Hugging Face},
  year = {2024}
}

@misc{hfLlama321bInstruct,
  author = {{Meta}},
  title = {Llama 3.2-1B Instruct Model Card},
  howpublished = {Hugging Face},
  year = {2024}
}

@misc{hfLlama323bInstruct,
  author = {{Meta}},
  title = {Llama 3.2-3B Instruct Model Card},
  howpublished = {Hugging Face},
  year = {2024}
}

@misc{hfMobileLLM1b,
  author = {{Meta}},
  title = {MobileLLM-1B Model Card},
  howpublished = {Hugging Face},
  year = {2024}
}

@misc{hfDeepSeekR1DistillQwen15b,
  author = {{DeepSeek}},
  title = {DeepSeek-R1-Qwen-1.5B Model Card},
  howpublished = {Hugging Face},
  year = {2024}
}

@misc{hfQwen25Zero5bInstruct,
  author = {{Alibaba}},
  title = {Qwen/Qwen2.5-0.5B-Instruct Model Card},
  howpublished = {Hugging Face},
  year = {2024}
}

@misc{hfQwen2515bInstruct,
  author = {{Alibaba}},
  title = {Qwen/Qwen2.5-1.5B-Instruct Model Card},
  howpublished = {Hugging Face},
  year = {2024}
}

@misc{hfQwen253bInstruct,
  author = {{Alibaba}},
  title = {Qwen/Qwen2.5-3B-Instruct Model Card},
  howpublished = {Hugging Face},
  year = {2024}
}

@misc{hfTinyLlama11bChat,
  author = {{TinyLlama}},
  title = {TinyLlama/TinyLlama-1.1B-Chat-v1.0 Model Card},
  howpublished = {Hugging Face},
  year = {2023}
}

@misc{hfZamba212bInstruct,
  author = {{Zyphra}},
  title = {Zyphra/Zamba2-1.2B-instruct Model Card},
  howpublished = {Hugging Face},
  year = {2024}
}

@misc{hfOlmo204251b,
  author = {{Allen Institute for AI}},
  title = {allenai/OLMo-2-0425-1B Model Card},
  howpublished = {Hugging Face},
  year = {2025}
}

@misc{hfDanube218bChat,
  author = {{H2O.ai}},
  title = {h2oai/h2o-danube2-1.8b-chat Model Card},
  howpublished = {Hugging Face},
  year = {2024}
}

@misc{hfDanube314bChat,
  author = {{H2O.ai}},
  title = {h2oai/h2o-danube3.1-4b-chat Model Card},
  howpublished = {Hugging Face},
  year = {2024}
}

@misc{hfSmolLM217bInstruct,
  author = {{Hugging Face}},
  title = {HuggingFaceTB/SmolLM2-1.7B-Instruct Model Card},
  howpublished = {Hugging Face},
}

@misc{hfSmolLM2135mInstruct,
  author = {{Hugging Face}},
  title = {HuggingFaceTB/SmolLM2-135M-Instruct Model Card},
  howpublished = {Hugging Face},
}

@misc{hfSmolLM2360mInstruct,
  author = {{Hugging Face}},
  title = {HuggingFaceTB/SmolLM2-360M-Instruct Model Card},
  howpublished = {Hugging Face},
}

@misc{hfStablelm216b,
  author = {{Stability AI}},
  title = {stabilityai/stablelm-2-1\_6b Model Card},
  howpublished = {Hugging Face},
  year = {2024}
}

@misc{hfDollyV23b,
  author = {{Databricks}},
  title = {databricks/dolly-v2-3b Model Card},
  howpublished = {Hugging Face},
  year = {2023}
}
